\documentclass{article}

\usepackage{hyperref}

\usepackage[accepted]{icml2022}

\usepackage[textsize=tiny]{todonotes}

\icmltitlerunning{Analysis of Adversarial Training at a Spectrum of Perturbations}
\usepackage{amsmath}
\usepackage{amsthm}
\usepackage{xcolor}
\usepackage{graphicx}
\usepackage{cite}
\usepackage{amsmath, amssymb}
\usepackage{mathtools}
\usepackage{soul}
\usepackage{algorithm}
\usepackage{subcaption}
\usepackage{mathrsfs}
\usepackage{todonotes}

\usepackage{amsmath,amsfonts,bm}

\def\eqref#1{equation~\ref{#1}}

\def\1{\bm{1}}

\DeclareMathAlphabet{\mathsfit}{\encodingdefault}{\sfdefault}{m}{sl}
\SetMathAlphabet{\mathsfit}{bold}{\encodingdefault}{\sfdefault}{bx}{n}

\newcommand{\E}{\mathbb{E}}

\newcommand{\D}{\mathcal{D}}

\newcommand{\Set}{\mathcal{S}}
\newcommand{\R}{\mathbb{R}}

\usepackage[autostyle]{csquotes}

\theoremstyle{plain}

\theoremstyle{definition}

\theoremstyle{remark}

\newcommand\reals{\mathbb{R}}

\definecolor{ed}{RGB}{225,0,0}
\definecolor{sd}{RGB}{0,0,225}
\definecolor{ks}{RGB}{225,100,0}

\newcommand{\bitem}{\begin{itemize}}
\newcommand{\eitem}{\end{itemize}}
\newcommand{\benum}{\begin{enumerate}}
\newcommand{\eenum}{\end{enumerate}}
\newcommand{\beq}{\begin{equation}}
\newcommand{\eeq}{\end{equation}}
\newcommand{\beqs}{\begin{equation*}}
\newcommand{\eeqs}{\end{equation*}}

\usepackage{amsfonts} 
\usepackage{colortbl}

\usepackage{diagbox}
\usepackage{float}

\begin{document}

\twocolumn[
\icmltitle{
Towards Alternative Techniques for Improving Adversarial Robustness: Analysis of Adversarial Training at a Spectrum of Perturbations 
}



\icmlsetsymbol{equal}{*}

\begin{icmlauthorlist}
\icmlauthor{Kaustubh Sridhar}{penn}
\icmlauthor{Souradeep Dutta}{penn}
\icmlauthor{Ramneet Kaur}{penn}
\icmlauthor{James Weimer}{penn}
\icmlauthor{Oleg Sokolsky}{penn}
\icmlauthor{Insup Lee}{penn}

\end{icmlauthorlist}

\icmlaffiliation{penn}{PRECISE Center, University of Pennsylvania}

\icmlcorrespondingauthor{Kaustubh Sridhar}{ksridhar@seas.upenn.edu}

\icmlkeywords{Adversarial Training, Robustness, Quantization}

\vskip 0.3in
]



\printAffiliationsAndNotice{}  

\begin{abstract}
Adversarial training (AT) and its variants have spearheaded progress in improving neural network robustness to adversarial perturbations and\\ common corruptions in the last few years. Algorithm design of AT and its variants are focused on training models at a specified perturbation strength $\epsilon$ and only using the feedback from the performance of that \emph{$\epsilon$-robust model} to improve the algorithm. In this work, we focus on models, trained on a spectrum of $\epsilon$ values. We analyze three perspectives: model performance, intermediate feature precision and convolution filter sensitivity. In each, we identify alternative improvements to AT that otherwise wouldn't have been apparent at a single $\epsilon$. Specifically, we find that for a PGD attack at some strength $\delta$, there is an AT model at some slightly larger strength $\epsilon$, but no greater, that generalizes best to it. Hence, we propose overdesigning for robustness where we suggest training models at an $\epsilon$ just above $\delta$. Second, we observe (across various $\epsilon$ values) that robustness is highly sensitive to the precision of intermediate features and particularly those after the first and second layer. Thus, we propose adding a simple quantization to defenses that improves accuracy on seen and unseen adaptive attacks. Third, we analyze convolution filters of each layer of models at increasing $\epsilon$ and notice that those of the first and second layer may be solely responsible for amplifying input perturbations. We present our findings and demonstrate our techniques through experiments with ResNet and WideResNet models on the CIFAR-10 and CIFAR-10-C datasets. \footnote{Code: \url{https://github.com/perturb-spectrum/analysis_and_methods}}
\end{abstract}

\section{Introduction} \label{sec:intro}

Adversarial training (AT) is currently the most effective method to improve the adversarial robustness of neural networks. AT and its variants have created robust models with state-of-the-art results against white-box attacks \citep{croce2020robustbench} without having to resort to obfuscated gradients \citep{athalye2018obfuscated}. Various studies on the effects of hyperparameters \citep{pang2020bag, sridhar2021improving, gowal2020uncovering, xie2019intriguing}, data augmentation \citep{rebuffi2021fixing}, unlabelled data \citep{carmon2019unlabeled, gowal2021improving, sehwag2021robust}, model size \citep{huang2021exploring, wu2021wider} and performance on unseen attacks \citep{stutz2020confidence, laidlaw2020perceptual} on AT have been conducted in the past. 

But each study has been restricted to analysis at a specified defense perturbation strength $\epsilon$ (usually 8/255 for $L_{\infty}$ adversarial robustness). For a fixed $\epsilon$-robust model, each of the studies above suggests an algorithm in addition to a combination of hyperparameters to improve robustness. By not observing the variation across $\epsilon$ values, current methods miss out on potential ideas that can aid robustness. In this work, we present three such ideas. These ideas employ feedback from three different sources.

First, based on the performance of models adversarially trained at various $\epsilon$'s, we find that there exists a defense perturbation strength $\epsilon$ (higher than the attack perturbation strength $\delta$) at which an AT model generalizes best to an attack. All models trained with strengths greater or lesser than this value are less effective defenses. We suggest \emph{overdesigning}, where for a given $\delta$, a defender must train models at increasing $\epsilon > \delta$ until 
the validation error starts deteriorating.  The recommendation would be to use the last model before degradation sets off, as the overdesigned defense (Section \ref{sec:gen}). In contrast, to the above adversarial robustness case are the more natural corruptions. Quite unexpectedly, we observe that lower $\epsilon$'s create better models for classifying images with common corruptions (Section \ref{sec:ood_all}) \& diagnose this issue as an effect of Pre-ReLU features. 

Second, we notice that the precision of intermediate features strongly influences robustness. We show that simple \emph{quantization} (especially after the first or second layer) increases robustness of models trained at various $\epsilon$'s to seen \& unseen adaptive attacks. 
This observation as well, is not obvious at a unit value of $\epsilon$, but can be seen to vary across the spectrum, for different values (Section \ref{sec:quant}).

Third, we show that the first and second layers' convolution filters are increasingly responsible for amplifying input perturbations in ResNet and WideResNet models for increasing $\epsilon$  (Section \ref{sec:change}). 

We demonstrate the above findings and the associated alternative techniques using ResNet18 and WideResNet-28-10 (with Swish activation functions) \citep{gowal2020uncovering} models on the CIFAR-10 and CIFAR-10-C \citep{hendrycks2019benchmarking} datasets. Our contributions are summarized below:
\vspace{-0.5em}
\begin{itemize}
    \item To the best of our knowledge, our work is the first to leverage information from adversarial trained $\epsilon$-robust models at increasing levels of $\epsilon$. Based on our observations, we suggest techniques to improve AT that is hard to fathom from a traditional analysis at one $\epsilon$.
    \item We propose an \emph{overdesigning} strategy that can consistently promise increased robustness to adversarial perturbations and demonstrate it with ResNet and WideResNet models on the CIFAR-10 dataset. 
    \item We propose \emph{quantization} of intermediate features that improves robustness for ResNet and WideResNet models on seen and unseen adaptive attacks (BPDA \citep{athalye2018obfuscated} and Transfer PGD \citep{croce2022evaluating}) on the CIFAR-10 dataset.
    \item We perform a study of convolution filters at increasing $\epsilon$ and identify the first and second layers of ResNet and WideResNet models, as sole suspects for input perturbation propagation. 
\end{itemize}
\vspace{-0.5em}


\section{Background} \label{sec:background}
\subsection{Adversarial Training}
Projected Gradient Descent based Adversarial training (PGD-AT) \citep{madry2018towards} solves a min-max optimization problem on a loss function $l$ as 
\begin{align}
    \min_{\theta} \; \E_{(x, y) \sim \D} \left[ \max_{x' \in \Set(x)} l(f_{\theta}(x'), y) \right]. \label{eq:AT}
\end{align}
Above, $\D$ represents the distribution over training data and $\Set(x)$ is the allowed set of perturbed examples around $x$, usually in an $L_p$ norm-bounded ball given by $\Set(x) = \{x' \; | \; \lvert\lvert x' - x  \rvert\rvert_p \leq \kappa\}$) where $\kappa$ is the perturbation strength. The inner maximization, which obtains the adversarial example, takes place with the projected gradient descent (PGD) attack and uses a step size $\alpha$
\begin{align}
    x^{i+1} = \Pi_{\Set(x)} \left(x^i + \alpha \; \text{sign}[ \nabla_{x^{i}} l(f_{\theta}(x^{i}), y) ] \right).
\end{align}
Where $x^0$ is chosen at random from within $\Set(x)$, $\Pi_{\Set(x)}(.)$ represents the projection into set $\Set(x)$, and the adversarial example is obtained after $N$ steps as $x' = x^N$. Here, $\kappa$ is set to $\epsilon/255$ where $\epsilon$ is an integer (usually, $8/255$ for $L_{\infty}$ perturbations). For a fair comparison, we set $\alpha = \frac{\epsilon}{8} \times \frac{2}{255}$ which takes the usual value of $\frac{2} {255}$ at $\epsilon = 8$ \citep{madry2018towards}. Many variants of PGD-AT have been proposed \citep{croce2022evaluating, zhang2019theoretically, carmon2019unlabeled, kang2021stable, huang2021exploring, wu2021wider, rade2021helper, pang2022robustness, sehwag2021robust}. In this work, for a rigorous study, we focus on the foundational method without the modifications. We refer to PGD-AT simply as AT throughout this paper.

\vspace{-0.5em}
\subsection{Notations}
Throughout this work, following \citet{madry2018towards} and for brevity, we refer to models trained as described above with perturbation strength $\epsilon/255$ as $\epsilon$-robust models. We denote an $N$ step PGD attack of strength $\delta$ as PGD$_{\delta}$-N.
\vspace{-0.5em}
\subsection{Adaptive Attacks}
When an improvement to AT is obtained through obfuscated gradients, adaptive attacks are necessary for evaluating the defense. Particularly, when a defense is not differentiable (such as with a scaled floor function in a quantization transformation), the Backward Pass Differentiable Approximation (BPDA) attack \citep{athalye2018obfuscated} is utilized where the non-differentiable defense is usually approximated with an identity function on the backward pass to compute gradients for an attack. Recent work \citep{croce2022evaluating} also recommends evaluating defenses that adapt models (including their intermediate features) against Transfer PGD, \textit{i.e.}, attacked images obtained from a PGD attack on just the model applied to the adapted model.

\vspace{-0.5em}
\section{Overdesigning for Robust Generalization} \label{sec:gen}
\begin{figure*}[t!]
\centering
\begin{subfigure}{.33\textwidth}
    \centering
    \includegraphics[width=\linewidth]{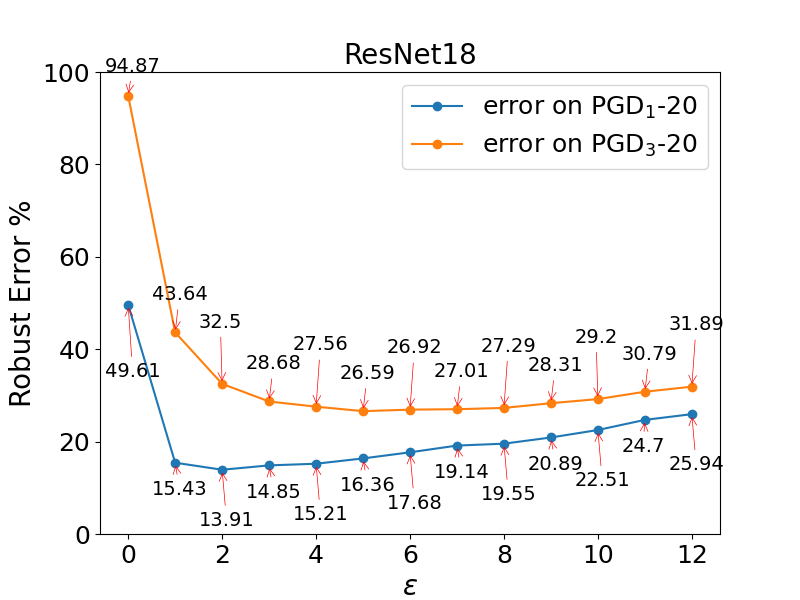}
\end{subfigure}
\begin{subfigure}{.33\textwidth}
    \centering
    \includegraphics[width=\linewidth]{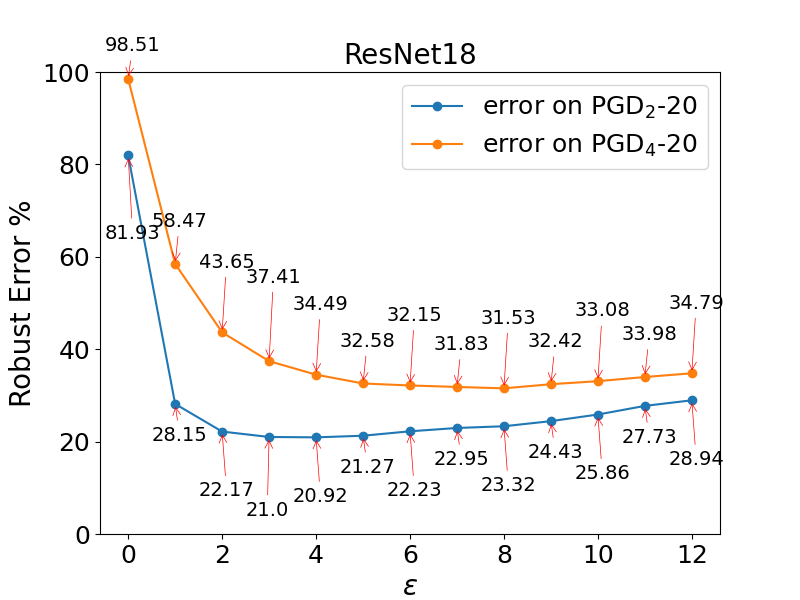}
\end{subfigure}
\begin{subfigure}{.33\textwidth}
    \centering
    \includegraphics[width=\linewidth]{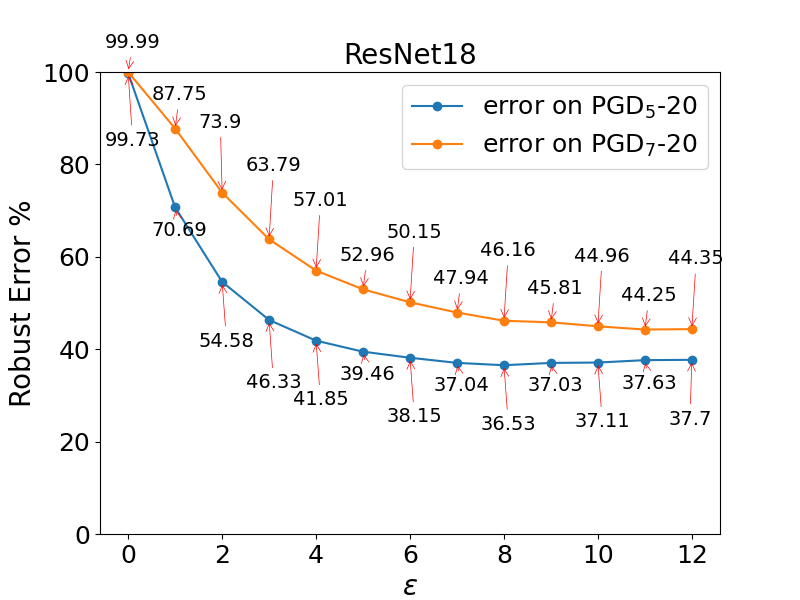}
\end{subfigure}
\begin{subfigure}{.33\textwidth}
    \centering
    \includegraphics[width=\linewidth]{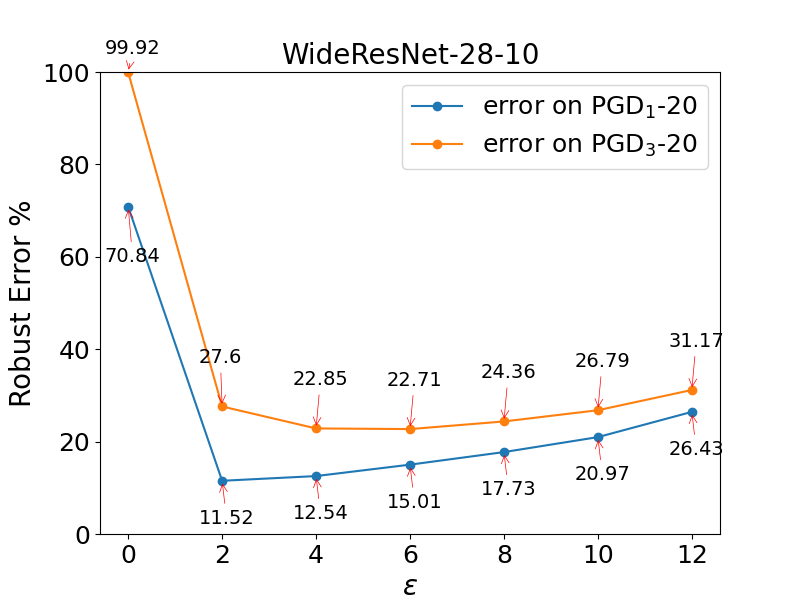}
\end{subfigure}
\begin{subfigure}{.33\textwidth}
    \centering
    \includegraphics[width=\linewidth]{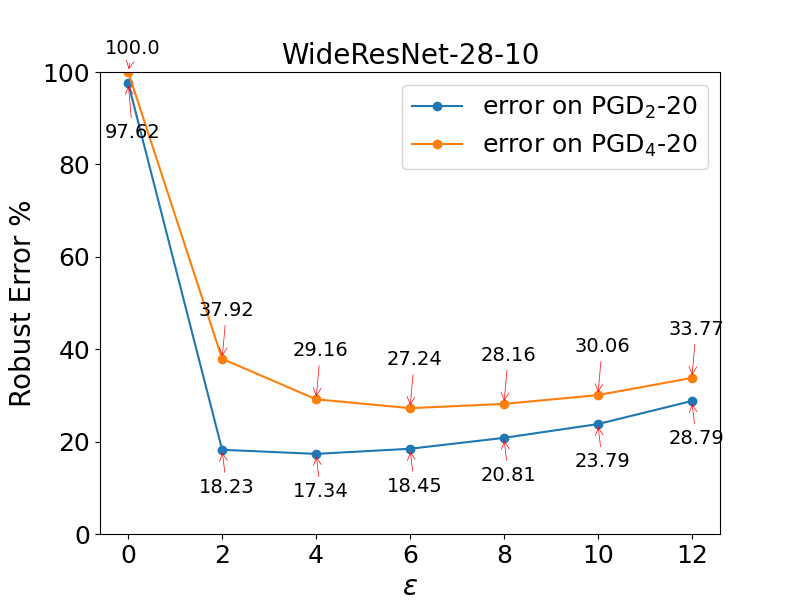}
\end{subfigure}
\begin{subfigure}{.33\textwidth}
    \centering
    \includegraphics[width=\linewidth]{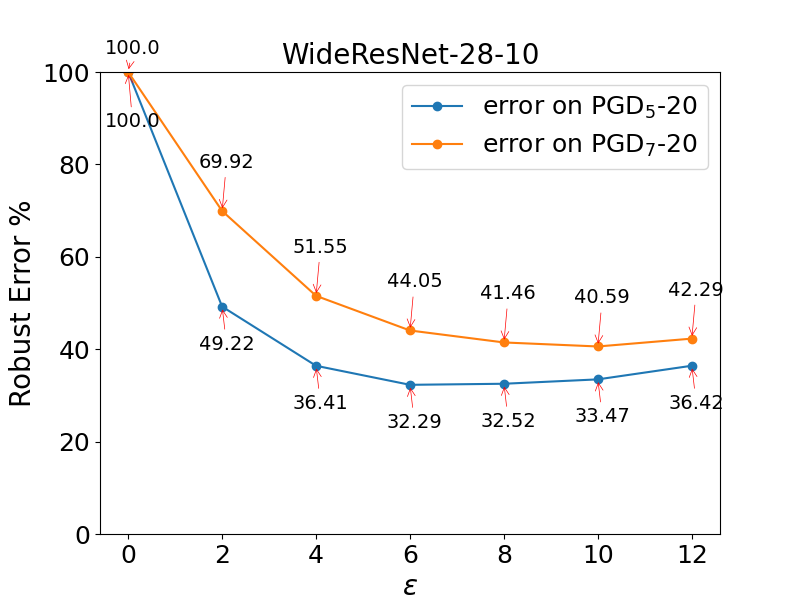}
\end{subfigure}
\caption{Robust error (= error
on PGD$_{\delta}$-20 on test set) of increasingly $\epsilon$-robust ResNet18 models for $\delta \in \{ 1, 3\}$ \emph{(top left)}, $\delta \in \{ 2, 4\}$ \emph{(top middle)}, $\delta \in \{ 5, 7\}$ \emph{(top right)}, and for WideResNet-28-10 models at $\delta \in \{ 1, 3\}$ \emph{(bottom left)}, $\delta \in \{2, 4\}$ \emph{(bottom middle)}, and $\delta \in \{ 5, 7\}$ \emph{(bottom right)}.}
\label{fig:g_err_mini}
\end{figure*}

\textbf{Motivation: }In a typical white-box attack case the attacker has full knowledge  of the internal model parameters and training hyperparameters. Post AT, the test and training perturbations respect a fixed $\epsilon$ contract on the attack strength. In this section we investigate : \textit{Does deviating from the norm, in terms of attack-strength  contracts, impact the overall robustness of the  model? } 

\textbf{Experiments: }We adversarial train multiple robust models with increasing values of the parameter $\epsilon$ and evaluate the models against PGD attacks of increasing strength $\delta$. The clean and robust accuracy along with the hyperparameters used, are given in Appendix \ref{app:individual}. The various $\epsilon$-robust models have similar clean accuracies, around $85\%$. We study the variation of errors (100 - accuracy) of the $\epsilon$-robust models on PGD$_{\delta}$-$20$ attacks for various $\delta > 0$. We show the plots of robust errors for $\epsilon$-robust models (from $\epsilon=0$ to $12$) with $\delta \in \{1, 3\}$, $\delta \in \{2, 4\}$ and $\delta \in \{5, 7\}$ for ResNet18 and WideResNet-28-10 models in Figure \ref{fig:g_err_mini}. 
 We plot the errors for other values of $\delta$ in Appendix \ref{app:change}.

\textbf{Discussions:}
In Figure \ref{fig:g_err_mini}, each curve represents the robust error against a PGD attack at a particular perturbation strength $\delta$. We notice that all four curves have a minima at $\epsilon > \delta$. For instance, in ResNet18 (Figure \ref{fig:g_err_mini} top row), the $2$-robust model generalizes best to the PGD$_{1}$-20 attack ($\epsilon_{*}-\delta=1$); the $4$-robust model generalizes best to the PGD$_{2}$-20 attack ($\epsilon_{*}-\delta=2$); the $5$-robust model generalizes best to the PGD$_{3}$-20 attack ($\epsilon_{*}-\delta=2$); the $8$-robust model generalizes best to the PGD$_{4}$-20 attack ($\epsilon_{*}-\delta=4$); the $9$-robust model generalizes best to the PGD$_{5}$-20 attack ($\epsilon_{*}-\delta=4$); the $11$-robust model generalizes best to the PGD$_{6}$-20 attack ($\epsilon_{*}-\delta=6$), and so on. Similar behaviour is seen for WideResNet-28-10 models (Figure \ref{fig:g_err_mini} bottom row).

This is surprising because, one would typically expect that a model trained with the largest $\epsilon$ should generalize best to all attacks of strength $\delta \leq \epsilon$. Especially when the clean accuracies are mostly similar. In contrast, the largest $\epsilon$-robust ResNet18 model above (\textit{i.e.}, the $12$-robust model) has up to an $8$\% larger error than the $\epsilon_{*}$-robust model. Further, from above, we observe that  $(\epsilon_{*}-\delta)$ increases with increasing $\delta$. This implies that eventually for very large $\delta$, the expected notion of the largest $\epsilon$-robust model generalizing best will be true. But at this large value of $\delta$, the perturbations would no longer be imperceptible to the human eye. 

Based on the aforementioned analysis, we suggest \emph{overdesigning} a defense by training models at $\epsilon > \delta$ until an increase in error is seen. For instance, in the case of ResNet18, the $\epsilon_{*}$-robust models for each $\delta$ listed above would be the right choice as the overdesigned model.

\vspace{-0.5em}
\section{Intermediate Feature Quantization}  \label{sec:quant}
\begin{table*}[t!]
\tiny
\centering
\begin{tabular}{p{0.35em}|c|c|c|c|c|c|c|c|c|c|c|c|c|c|c|c|c|c}
$\delta$ & Quant- & \multicolumn{4}{c|}{Transf. PGD$_{\delta}$-20 on ResNet18} & \multicolumn{4}{c|}{BPDA$_{\delta}$-20 on ResNet18} & Quant- & \multicolumn{4}{c|}{Transf. PGD$_{\delta}$-20 on WRN-28-10} & \multicolumn{4}{c}{BPDA$_{\delta}$ on WRN-28-10} \\
\cline{3-10} \cline{12-19}
 & ization & \multicolumn{4}{c|}{$\epsilon$} & \multicolumn{4}{c|}{$\epsilon$} & ization & \multicolumn{4}{c|}{$\epsilon$} & \multicolumn{4}{c}{$\epsilon$} \\
\cline{3-10} \cline{12-19}
 & after & $2$ & $4$ & $8$ & $12$ & $2$ & $4$ & $8$ & $12$ & after & $2$ & $4$ & $8$ & $12$ & $2$ & $4$ & $8$ & $12$\\
 & layer: & & & & & & & & & layer: & & & & & & & & \\
\hline

2 & none & \cellcolor{gray!10}77.85 & \cellcolor{gray!10}79.08 & \cellcolor{gray!10}76.69 & \cellcolor{gray!10}71.01 & \cellcolor{gray!10}77.85 & \cellcolor{gray!10}79.08 & \cellcolor{gray!10}76.69 & \cellcolor{gray!10}71.01 & none & \cellcolor{gray!10}81.77 & \cellcolor{gray!10}82.67 & \cellcolor{gray!10}79.19 & \cellcolor{gray!10}71.2 & \cellcolor{gray!10}81.77 & \cellcolor{gray!10}82.67 & \cellcolor{gray!10}79.19 & \cellcolor{gray!10}71.2 \\
 & conv0 & \cellcolor{blue!40}79.03 & \cellcolor{blue!32}79.42 & \cellcolor{red!6}75.88 & \cellcolor{red!6}68.9 & \cellcolor{red!6}77.29 & \cellcolor{red!6}78.29 & \cellcolor{red!6}75.25 & \cellcolor{red!6}68.41 & init conv & \cellcolor{red!6}81.63 & \cellcolor{blue!40}82.76 & \cellcolor{red!6}79.06 & \cellcolor{red!6}70.9 & \cellcolor{red!6}81.62 & \cellcolor{blue!32}82.76 & \cellcolor{red!6}79.13 & \cellcolor{red!6}70.9 \\
 & layer1 & \cellcolor{blue!19}78.43 & \cellcolor{blue!21}79.31 & \cellcolor{red!6}76.46 & \cellcolor{red!6}70.3 & \cellcolor{blue!40}78.0 & \cellcolor{red!6}79.07 & \cellcolor{red!6}76.38 & \cellcolor{red!6}70.28 & layer[0] & \cellcolor{blue!40}81.85 & \cellcolor{red!6}82.6 & \cellcolor{red!6}78.92 & \cellcolor{red!6}70.92 & \cellcolor{blue!40}81.85 & \cellcolor{red!6}82.62 & \cellcolor{red!6}78.93 & \cellcolor{red!6}70.98 \\
 & layer2 & \cellcolor{blue!7}78.01 & \cellcolor{blue!21}79.31 & \cellcolor{blue!26}76.71 & \cellcolor{red!6}70.15 & 77.85 & \cellcolor{blue!8}79.15 & \cellcolor{red!6}76.53 & \cellcolor{red!6}70.14 & layer[1] & \cellcolor{red!6}81.72 & \cellcolor{blue!40}82.76 & \cellcolor{red!6}79.01 & \cellcolor{red!6}70.46 & \cellcolor{red!6}81.74 & \cellcolor{blue!25}82.74 & \cellcolor{red!6}79.02 & \cellcolor{red!6}70.5 \\
 & layer3 & \cellcolor{red!6}77.84 & \cellcolor{blue!40}79.5 & \cellcolor{blue!40}76.72 & \cellcolor{red!6}70.97 & \cellcolor{blue!7}77.86 & \cellcolor{blue!40}79.42 & \cellcolor{blue!7}76.71 & \cellcolor{red!6}70.95 & layer[2] & \cellcolor{red!6}81.68 & \cellcolor{blue!31}82.74 & \cellcolor{red!6}79.0 & \cellcolor{red!6}70.78 & \cellcolor{red!6}81.74 & \cellcolor{blue!40}82.78 & \cellcolor{red!6}79.11 & \cellcolor{red!6}70.9 \\
 & layer4 & \cellcolor{blue!7}77.92 & \cellcolor{blue!7}79.14 & 76.69 & \cellcolor{red!6}70.58 & \cellcolor{blue!21}77.93 & \cellcolor{blue!14}79.2 & \cellcolor{blue!40}76.91 & \cellcolor{red!6}70.79 &  &  &  &  &  &  &  &  &  \\
\hline
4 & none & \cellcolor{gray!10}56.39 & \cellcolor{gray!10}65.46 & \cellcolor{gray!10}68.5 & \cellcolor{gray!10}65.2 & \cellcolor{gray!10}56.39 & \cellcolor{gray!10}65.46 & \cellcolor{gray!10}68.5 & \cellcolor{gray!10}65.2 & none & \cellcolor{gray!10}62.12 & \cellcolor{gray!10}70.87 & \cellcolor{gray!10}71.81 & \cellcolor{gray!10}66.2 & \cellcolor{gray!10}62.12 & \cellcolor{gray!10}70.87 & \cellcolor{gray!10}71.81 & \cellcolor{gray!10}66.2 \\
 & conv0 & \cellcolor{blue!40}61.4 & \cellcolor{blue!40}67.61 & \cellcolor{red!6}68.23 & \cellcolor{red!6}63.99 & \cellcolor{blue!40}57.47 & \cellcolor{red!6}65.27 & \cellcolor{red!6}67.18 & \cellcolor{red!6}62.92 & init conv & \cellcolor{red!6}61.86 & \cellcolor{red!6}70.83 & \cellcolor{red!6}71.53 & \cellcolor{red!6}65.87 & \cellcolor{red!6}61.73 & \cellcolor{red!6}70.8 & \cellcolor{red!6}71.53 & \cellcolor{red!6}65.81 \\
 & layer1 & \cellcolor{blue!9}57.57 & \cellcolor{blue!14}66.23 & \cellcolor{red!6}68.28 & \cellcolor{red!6}64.95 & \cellcolor{blue!18}56.88 & \cellcolor{blue!26}65.72 & \cellcolor{red!6}68.12 & \cellcolor{red!6}64.87 & layer[0] & \cellcolor{blue!40}62.43 & \cellcolor{blue!24}70.9 & \cellcolor{red!6}71.78 & \cellcolor{red!6}65.97 & \cellcolor{blue!40}62.39 & \cellcolor{blue!18}70.95 & \cellcolor{blue!7}71.82 & \cellcolor{red!6}66.06 \\
 & layer2 & \cellcolor{blue!7}57.12 & \cellcolor{blue!11}66.09 & \cellcolor{blue!7}68.52 & \cellcolor{red!6}64.78 & \cellcolor{blue!7}56.53 & \cellcolor{blue!40}65.85 & \cellcolor{red!6}68.23 & \cellcolor{red!6}64.85 & layer[1] & 62.12 & \cellcolor{blue!40}70.92 & \cellcolor{red!6}71.74 & \cellcolor{red!6}65.82 & \cellcolor{blue!7}62.14 & \cellcolor{blue!7}70.9 & \cellcolor{blue!7}71.82 & \cellcolor{red!6}65.82 \\
 & layer3 & \cellcolor{blue!7}56.73 & \cellcolor{blue!7}65.84 & \cellcolor{blue!40}68.83 & \cellcolor{blue!40}65.37 & \cellcolor{blue!7}56.59 & \cellcolor{blue!36}65.82 & \cellcolor{blue!40}68.54 & \cellcolor{blue!31}65.55 & layer[2] & \cellcolor{blue!7}62.15 & \cellcolor{blue!24}70.9 & \cellcolor{red!6}71.77 & \cellcolor{red!6}65.7 & \cellcolor{blue!7}62.17 & \cellcolor{blue!40}71.04 & \cellcolor{blue!40}71.9 & \cellcolor{red!6}65.88 \\
 & layer4 & \cellcolor{blue!7}56.54 & \cellcolor{blue!7}65.52 & \cellcolor{red!6}68.25 & 65.2 & \cellcolor{blue!7}56.49 & \cellcolor{blue!15}65.61 & \cellcolor{blue!10}68.51 & \cellcolor{blue!40}65.65 &  &  &  &  &  &  &  &  &  \\
\hline
6 & none & \cellcolor{gray!10}35.31 & \cellcolor{gray!10}50.31 & \cellcolor{gray!10}58.6 & \cellcolor{gray!10}58.97 & \cellcolor{gray!10}35.31 & \cellcolor{gray!10}50.31 & \cellcolor{gray!10}58.6 & \cellcolor{gray!10}58.97 & none & \cellcolor{gray!10}39.5 & \cellcolor{gray!10}56.25 & \cellcolor{gray!10}62.91 & \cellcolor{gray!10}60.54 & \cellcolor{gray!10}39.5 & \cellcolor{gray!10}56.25 & \cellcolor{gray!10}62.91 & \cellcolor{gray!10}60.54 \\
 & conv0 & \cellcolor{blue!40}41.13 & \cellcolor{blue!40}54.46 & \cellcolor{blue!40}60.0 & \cellcolor{red!6}58.61 & \cellcolor{blue!40}36.85 & \cellcolor{blue!40}51.45 & \cellcolor{red!6}58.22 & \cellcolor{red!6}57.49 & init conv & \cellcolor{red!6}39.4 & \cellcolor{blue!7}56.28 & \cellcolor{blue!8}62.92 & \cellcolor{blue!40}60.61 & \cellcolor{red!6}39.3 & \cellcolor{red!6}56.19 & \cellcolor{blue!7}62.95 & \cellcolor{blue!32}60.72 \\
 & layer1 & \cellcolor{blue!7}36.22 & \cellcolor{blue!11}51.55 & \cellcolor{blue!7}58.85 & \cellcolor{blue!7}59.06 & \cellcolor{blue!7}35.43 & \cellcolor{blue!21}50.91 & \cellcolor{blue!9}58.7 & \cellcolor{red!6}58.83 & layer[0] & \cellcolor{blue!40}40.42 & \cellcolor{blue!40}56.73 & \cellcolor{red!6}62.88 & \cellcolor{red!6}60.48 & \cellcolor{blue!40}40.28 & \cellcolor{blue!40}56.61 & \cellcolor{blue!12}63.01 & \cellcolor{blue!9}60.59 \\
 & layer2 & \cellcolor{blue!7}36.04 & \cellcolor{blue!7}50.99 & \cellcolor{blue!7}58.83 & \cellcolor{blue!11}59.12 & \cellcolor{blue!7}35.61 & \cellcolor{blue!17}50.8 & \cellcolor{red!6}58.59 & \cellcolor{blue!11}59.18 & layer[1] & \cellcolor{blue!10}39.75 & \cellcolor{blue!14}56.42 & \cellcolor{blue!40}62.96 & \cellcolor{red!6}60.26 & \cellcolor{blue!11}39.73 & \cellcolor{blue!8}56.33 & \cellcolor{blue!8}62.98 & \cellcolor{red!6}60.44 \\
 & layer3 & \cellcolor{blue!7}35.38 & \cellcolor{blue!7}50.6 & \cellcolor{blue!11}58.99 & \cellcolor{blue!40}59.47 & \cellcolor{red!6}35.19 & \cellcolor{blue!12}50.68 & \cellcolor{blue!27}58.9 & \cellcolor{blue!40}59.68 & layer[2] & \cellcolor{blue!7}39.54 & \cellcolor{blue!7}56.26 & \cellcolor{blue!23}62.94 & \cellcolor{red!6}60.45 & \cellcolor{blue!7}39.61 & \cellcolor{blue!7}56.32 & \cellcolor{blue!40}63.23 & \cellcolor{blue!40}60.76 \\
 & layer4 & \cellcolor{red!6}35.29 & \cellcolor{blue!7}50.39 & \cellcolor{red!6}58.54 & \cellcolor{red!6}58.82 & \cellcolor{blue!7}35.32 & \cellcolor{blue!7}50.41 & \cellcolor{blue!40}59.04 & \cellcolor{blue!38}59.65 &  &  &  &  &  &  &  &  &  \\
\hline
8 & none & \cellcolor{gray!10}19.37 & \cellcolor{gray!10}36.2 & \cellcolor{gray!10}48.77 & \cellcolor{gray!10}52.53 & \cellcolor{gray!10}19.37 & \cellcolor{gray!10}36.2 & \cellcolor{gray!10}48.77 & \cellcolor{gray!10}52.53 & none & \cellcolor{gray!10}22.08 & \cellcolor{gray!10}41.2 & \cellcolor{gray!10}53.72 & \cellcolor{gray!10}54.67 & \cellcolor{gray!10}22.08 & \cellcolor{gray!10}41.2 & \cellcolor{gray!10}53.72 & \cellcolor{gray!10}54.67 \\
 & conv0 & \cellcolor{blue!40}24.48 & \cellcolor{blue!40}40.58 & \cellcolor{blue!40}51.0 & \cellcolor{blue!40}52.87 & \cellcolor{blue!40}20.78 & \cellcolor{blue!40}37.4 & \cellcolor{red!6}48.74 & \cellcolor{red!6}51.15 & init conv & \cellcolor{red!6}21.88 & \cellcolor{blue!7}41.26 & \cellcolor{blue!14}53.8 & \cellcolor{red!6}54.34 & \cellcolor{red!6}21.88 & \cellcolor{red!6}41.17 & \cellcolor{blue!11}53.84 & \cellcolor{red!6}54.46 \\
 & layer1 & \cellcolor{blue!7}19.82 & \cellcolor{blue!10}37.31 & \cellcolor{blue!10}49.36 & \cellcolor{red!6}52.45 & \cellcolor{blue!7}19.38 & \cellcolor{blue!20}36.82 & \cellcolor{blue!25}49.28 & \cellcolor{red!6}52.41 & layer[0] & \cellcolor{blue!40}22.75 & \cellcolor{blue!40}41.74 & \cellcolor{blue!36}53.92 & \cellcolor{red!6}54.42 & \cellcolor{blue!40}22.56 & \cellcolor{blue!40}41.69 & \cellcolor{blue!40}54.13 & \cellcolor{blue!7}54.69 \\
 & layer2 & \cellcolor{blue!7}19.88 & \cellcolor{blue!7}36.93 & \cellcolor{blue!11}49.43 & \cellcolor{blue!24}52.74 & \cellcolor{blue!7}19.61 & \cellcolor{blue!13}36.6 & \cellcolor{blue!27}49.33 & \cellcolor{blue!16}52.9 & layer[1] & \cellcolor{blue!9}22.24 & \cellcolor{blue!14}41.39 & \cellcolor{blue!40}53.94 & \cellcolor{red!6}54.49 & \cellcolor{blue!7}22.12 & \cellcolor{blue!18}41.43 & \cellcolor{blue!34}54.07 & \cellcolor{red!6}54.57 \\
 & layer3 & \cellcolor{blue!7}19.42 & \cellcolor{blue!7}36.42 & \cellcolor{blue!14}49.56 & \cellcolor{blue!38}52.86 & \cellcolor{blue!7}19.38 & \cellcolor{blue!7}36.32 & \cellcolor{blue!40}49.58 & \cellcolor{blue!23}53.06 & layer[2] & \cellcolor{red!6}22.01 & \cellcolor{blue!7}41.21 & \cellcolor{blue!9}53.77 & \cellcolor{red!6}54.62 & \cellcolor{blue!7}22.14 & \cellcolor{blue!10}41.33 & \cellcolor{blue!36}54.09 & \cellcolor{blue!40}54.99 \\
 & layer4 & \cellcolor{blue!7}19.38 & \cellcolor{red!6}36.19 & \cellcolor{blue!7}48.84 & \cellcolor{red!6}52.41 & \cellcolor{red!6}19.3 & \cellcolor{blue!7}36.29 & \cellcolor{blue!39}49.56 & \cellcolor{blue!40}53.42 &  &  &  &  &  &  &  &  &  \\
\hline
10 & none & \cellcolor{gray!10}9.63 & \cellcolor{gray!10}24.41 & \cellcolor{gray!10}39.64 & \cellcolor{gray!10}45.67 & \cellcolor{gray!10}9.63 & \cellcolor{gray!10}24.41 & \cellcolor{gray!10}39.64 & \cellcolor{gray!10}45.67 & none & \cellcolor{gray!10}11.15 & \cellcolor{gray!10}28.19 & \cellcolor{gray!10}44.65 & \cellcolor{gray!10}48.12 & \cellcolor{gray!10}11.15 & \cellcolor{gray!10}28.19 & \cellcolor{gray!10}44.65 & \cellcolor{gray!10}48.12 \\
 & conv0 & \cellcolor{blue!40}13.15 & \cellcolor{blue!40}28.58 & \cellcolor{blue!40}41.99 & \cellcolor{blue!40}46.93 & \cellcolor{blue!40}10.67 & \cellcolor{blue!40}25.17 & \cellcolor{red!6}39.29 & \cellcolor{red!6}44.87 & init conv & \cellcolor{red!6}10.89 & \cellcolor{blue!7}28.28 & \cellcolor{red!6}44.58 & \cellcolor{blue!16}48.34 & \cellcolor{red!6}11.11 & \cellcolor{red!6}28.13 & \cellcolor{red!6}44.64 & \cellcolor{blue!13}48.34 \\
 & layer1 & \cellcolor{blue!7}10.01 & \cellcolor{blue!9}25.38 & \cellcolor{blue!11}40.31 & \cellcolor{blue!7}45.87 & \cellcolor{blue!7}9.68 & \cellcolor{blue!31}25.01 & \cellcolor{blue!26}40.21 & \cellcolor{blue!7}45.73 & layer[0] & \cellcolor{blue!40}11.69 & \cellcolor{blue!40}28.84 & \cellcolor{blue!40}45.02 & \cellcolor{blue!24}48.45 & \cellcolor{blue!40}11.44 & \cellcolor{blue!40}28.55 & \cellcolor{blue!40}45.0 & \cellcolor{blue!36}48.7 \\
 & layer2 & \cellcolor{blue!7}9.9 & \cellcolor{blue!7}24.99 & \cellcolor{blue!13}40.46 & \cellcolor{blue!7}45.91 & \cellcolor{blue!7}9.75 & \cellcolor{blue!20}24.8 & \cellcolor{blue!29}40.26 & \cellcolor{blue!20}46.34 & layer[1] & \cellcolor{blue!7}11.23 & \cellcolor{blue!17}28.47 & \cellcolor{blue!18}44.82 & \cellcolor{blue!40}48.66 & \cellcolor{red!6}11.02 & \cellcolor{blue!14}28.32 & \cellcolor{blue!24}44.86 & \cellcolor{blue!40}48.76 \\
 & layer3 & \cellcolor{red!6}9.52 & \cellcolor{blue!7}24.59 & \cellcolor{blue!8}40.15 & \cellcolor{blue!7}45.91 & \cellcolor{red!6}9.25 & \cellcolor{blue!7}24.53 & \cellcolor{blue!30}40.28 & \cellcolor{blue!13}46.1 & layer[2] & 11.15 & \cellcolor{blue!9}28.34 & \cellcolor{red!6}44.61 & \cellcolor{blue!7}48.2 & \cellcolor{red!6}11.09 & \cellcolor{blue!20}28.37 & \cellcolor{blue!21}44.84 & \cellcolor{blue!32}48.64 \\
 & layer4 & \cellcolor{blue!7}9.69 & \cellcolor{blue!7}24.42 & \cellcolor{blue!7}39.66 & \cellcolor{red!6}45.55 & \cellcolor{red!6}9.52 & \cellcolor{blue!7}24.55 & \cellcolor{blue!40}40.49 & \cellcolor{blue!40}46.96 &  &  &  &  &  &  &  &  &  \\
\hline
12 & none & \cellcolor{gray!10}4.28 & \cellcolor{gray!10}15.48 & \cellcolor{gray!10}31.17 & \cellcolor{gray!10}39.27 & \cellcolor{gray!10}4.28 & \cellcolor{gray!10}15.48 & \cellcolor{gray!10}31.17 & \cellcolor{gray!10}39.27 & none & \cellcolor{gray!10}5.18 & \cellcolor{gray!10}17.54 & \cellcolor{gray!10}35.93 & \cellcolor{gray!10}42.09 & \cellcolor{gray!10}5.18 & \cellcolor{gray!10}17.54 & \cellcolor{gray!10}35.93 & \cellcolor{gray!10}42.09 \\
 & conv0 & \cellcolor{blue!40}6.51 & \cellcolor{blue!40}18.79 & \cellcolor{blue!40}33.62 & \cellcolor{blue!40}40.74 & \cellcolor{blue!40}5.28 & \cellcolor{blue!40}16.1 & \cellcolor{red!6}30.85 & \cellcolor{red!6}38.76 & init conv & \cellcolor{red!6}5.08 & \cellcolor{blue!7}17.62 & \cellcolor{red!6}35.79 & \cellcolor{red!6}42.06 & \cellcolor{red!6}5.15 & \cellcolor{blue!7}17.66 & \cellcolor{red!6}35.84 & \cellcolor{blue!7}42.2 \\
 & layer1 & \cellcolor{blue!7}4.42 & \cellcolor{blue!7}16.13 & \cellcolor{blue!9}31.77 & \cellcolor{blue!7}39.54 & \cellcolor{blue!7}4.29 & \cellcolor{blue!7}15.6 & \cellcolor{blue!19}31.56 & \cellcolor{blue!10}39.57 & layer[0] & \cellcolor{blue!40}5.35 & \cellcolor{blue!40}18.19 & \cellcolor{blue!24}36.18 & \cellcolor{blue!16}42.21 & \cellcolor{blue!40}5.35 & \cellcolor{blue!40}18.44 & \cellcolor{blue!27}36.31 & \cellcolor{blue!23}42.42 \\
 & layer2 & \cellcolor{blue!7}4.5 & \cellcolor{blue!7}16.11 & \cellcolor{blue!14}32.05 & \cellcolor{blue!17}39.91 & \cellcolor{blue!7}4.37 & \cellcolor{blue!19}15.79 & \cellcolor{blue!40}31.95 & \cellcolor{blue!40}40.43 & layer[1] & \cellcolor{blue!16}5.25 & \cellcolor{blue!19}17.85 & \cellcolor{blue!40}36.34 & \cellcolor{blue!40}42.38 & \cellcolor{red!6}5.14 & \cellcolor{blue!15}17.9 & \cellcolor{blue!37}36.45 & \cellcolor{blue!35}42.59 \\
 & layer3 & \cellcolor{blue!7}4.3 & \cellcolor{blue!7}15.62 & \cellcolor{blue!8}31.72 & \cellcolor{blue!7}39.52 & \cellcolor{blue!7}4.29 & \cellcolor{blue!7}15.52 & \cellcolor{blue!37}31.91 & \cellcolor{blue!13}39.66 & layer[2] & \cellcolor{red!6}5.15 & \cellcolor{blue!7}17.63 & \cellcolor{blue!28}36.22 & \cellcolor{blue!23}42.26 & \cellcolor{red!6}5.03 & \cellcolor{blue!7}17.72 & \cellcolor{blue!40}36.49 & \cellcolor{blue!40}42.65 \\
 & layer4 & \cellcolor{blue!7}4.3 & \cellcolor{red!6}15.45 & \cellcolor{red!6}31.16 & \cellcolor{red!6}39.02 & \cellcolor{blue!7}4.34 & \cellcolor{blue!9}15.63 & \cellcolor{blue!30}31.76 & \cellcolor{blue!37}40.36 &  &  &  &  &  &  &  &  &  \\
\hline

\end{tabular}
\caption{Evaluation of quantization on intermediate features of $\epsilon$-robust ResNet18 and WideResNet-28-10 models on the Transfer PGD$_{\delta}$-20 and BPDA$_{\delta}$ attacks for various $\epsilon$ and $\delta$. The intensity of colors in the shaded cells is in proportion to the magnitude of increase over the non-quantized accuracies.}
\label{tab:mini_quant}
\end{table*}

\textbf{Motivation: }Quantization offers a simple alternative to induce robustness to input perturbations. Simply on account of the reduced sensitivity to perturbations away from quantization boundaries. There exists some prior work on this in literature, \citep{buckman2018thermometer, guo2017countering} but have been unable to withstand adaptive attacks like BPDA \citep{athalye2018obfuscated}. This seemingly arbitrary defense does seem to have some impact on the robustness of the overall computation. We attempt to study this in more detail here. In particular, we ask the question - \textit{what changes in the intermediate features affect robust accuracy while minimally changing clean accuracy? } 

\textbf{Experiments: }We operate at the level of convolutional blocks. We quantize intermediate features after one of the five convolutional blocks  in various adversarially trained ResNet18 models, and one of the four blocks in WideResNet-28-10 models. 
The quantization is performed with a simple element-wise scaled floor function on input tensor $x \in \R^{C\times W \times H}$ given as $\lfloor \beta x_{ijk} \rfloor / \beta$, for some scalar $\beta$. Notice that the above operation can be done for any intermediate neuron in the DNN, and can be prohibitively large to explore in an individual basis. But, a useful abstraction is to work at the level of layers of the DNN. We evaluate feature-quantization on two adaptive attacks, namely, BPDA$_{\delta}$ from \citet{athalye2018obfuscated} and Transfer PDG$_{\delta}$-20 from \citet{croce2022evaluating} for various $\delta$'s and show some sample results in Table \ref{tab:mini_quant}. The complete set of results for $\epsilon \in [0, 12]$ and $\delta \in \{0, 2, 4, 6, 8\}$, clean accuracies after quantization and ablations for various $\beta$ values can be found in Appendix \ref{app:quant}.

\textbf{Discussions: }In Table \ref{tab:mini_quant}, the cells shaded blue represent an increase over the unquantized model (gray cells) while, cells shaded red denote a loss in accuracy.  The dark blue shaded lines in Table \ref{tab:mini_quant} correspond to accuracies obtained with quantization after `conv0' for ResNet18 and after `layer[0]' for WideResNet-28-10. They denote significant increase in accuracy with quantization of intermediate features after the first layer in ResNet18 and after the second layer in WideResNet-28-10. Whereas, the red shaded cells on the top right show a loss in robustness on the weak attack $\delta=2$ for high strength ($\epsilon=8, ..., 12$) models. This can be understood as the expected (but minimal) loss in accuracy with a drop in resolution for features of attacked images that are almost the same as clean images.

Unexpectedly, we also notice that the shaded blue values span across seen ($\epsilon \geq \delta$) and unseen attacks ($\epsilon < \delta$). Further, with individual columns of $\epsilon$-robust models like the $\epsilon=12$ column under the transfer PGD attack on WRN-28-10 and the $\epsilon=2$ column under the BPDA attack on WRN-28-10, it would have been easy to reject this technique as unsuitable. Yet, with an evaluation spanning across increasing values of $\epsilon$, we notice that broad improvements are attained in both seen and unseen attacks, other columns of Table \ref{tab:mini_quant}. Our evaluation confirms that quantization can be applied to any model to provide both seen and unseen robustness without explicit calibration to unseen attacks.



\vspace{-0.5em}
\section{AT and Norm of CNN Kernels} \label{sec:change}

\begin{figure*}[t]
    \centering
    \begin{subfigure}{0.37\textwidth}
        \includegraphics[width=\linewidth]{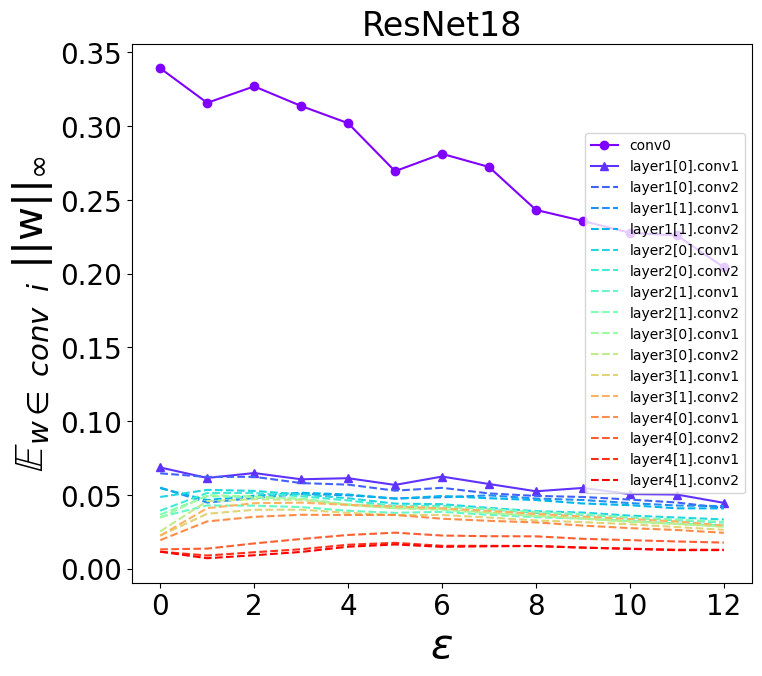}
    \end{subfigure}
    \begin{subfigure}{0.44\textwidth}
        \includegraphics[width=\linewidth]{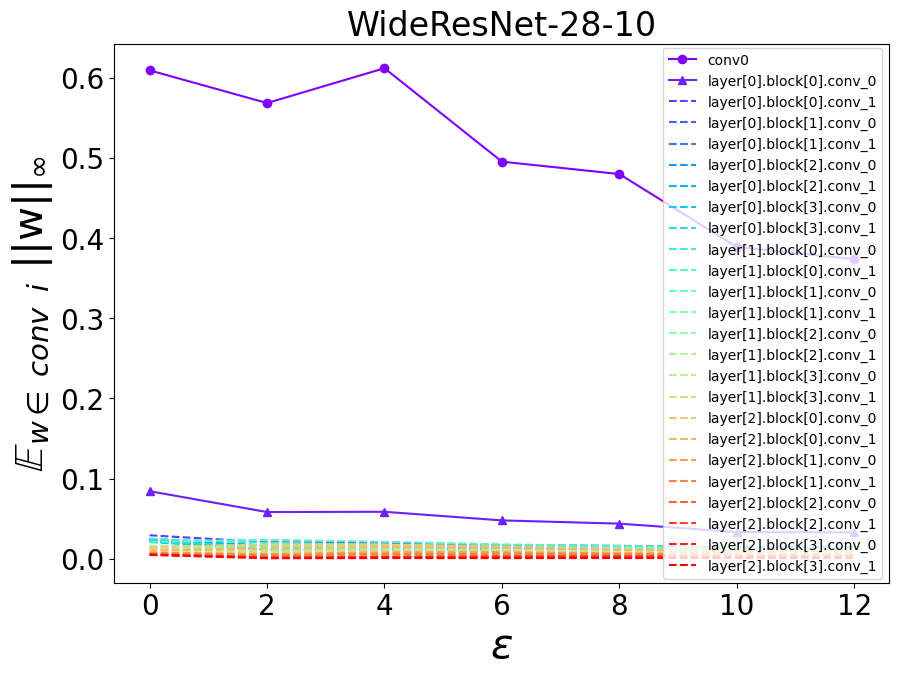}
    \end{subfigure}
    \vspace{-1em}
    \caption{Plot of maximum of $L_{\infty}$ norms of filters in various layers for increasingly $\epsilon$-robust ResNet18 \emph{(left)} and WideResNet-28-10 \emph{(right)} models.}
    \label{fig:hist}
\end{figure*}
\begin{table*}[t!]
    \scriptsize 
    \centering
    \begin{tabular}{l|c|c|c|c|c|c|c|c|c|c|c|c|c}

Model & non-robust & 1 & 2 & 3 & 4 & 5 & 6 & 7 & 8 & 9 & 10 & 11 & 12 \\
\hline
ResNet18 & 66.0 & \underline{81.8} & \textbf{81.97} & 81.28 & 80.19 & 79.07 & 77.83 & 76.33 & 75.46 & 74.14 & 72.62 & 70.82 & 69.37 \\
WideResNet-28-10 & 58.48 &- & \textbf{84.0} &- & \underline{82.49} &- & 79.74 &- & 76.68 &- & 73.53 &- & 68.48 \\
\hline

    \end{tabular}
    \caption{Average accuracy of $\epsilon$-robust ResNet18 and WideResNet-28-10 models (for various $\epsilon$) on all common corruptions in CIFAR-10-C dataset. The bold and underlined values denote the largest and second largest value in each row.}
    \label{tab:mini_corruptions}
    \vspace{-0.5em}
\end{table*}

\textbf{Motivation: }In this section  we attempt to analyze the variation in the norm of the kernel weights for different layers across increasingly robust models. The kernel weights are square matrices of shape $3 \times 3$. Each convolutional operation in a CNN computes an output of the form $y = w^Tv$, where $\{w, v\} \in \reals^d$ and $d = 9$. $v$ represents the part of the input the kernel gets applied to. Then by Cauchy-Schwartz inequality $||y|| = || w^Tv|| \leq ||w|| ||v|| $. The norm $||w||_\infty = max_i |w_i|$ serves as the $L_\infty$  Lipschitz constant for $y$ w.r.t to the input $v$. Thus, we analyze the $L_{\infty}$ norm of kernel weights to study the sensitivity of convolutional layers to $L_\infty$ norm bounded perturbations. 

\textbf{Experiments: }We analyze the average and worst case performance for each layer in ResNet18 and WideResNet-28-10 models, by analyzing the average filter sensitivity and the filter that is most sensitive to input perturbations respectively. These can be measured by the average and maximum over the filter norms in each layer. That is, for a particular convolution layer $i$, we compute $\E_{w \in \text{ conv i}} ||w||_{\infty}$ and $\max_{w \in \text{ conv i}} ||w||_{\infty}$. We plot the former for each layer of increasingly $\epsilon$-robust ResNet18 and WideResNet-28-10 models in Figure \ref{fig:hist} and the latter in Appendix \ref{app:change}. 

\begin{figure}[t]
\centering
\begin{subfigure}{.45\linewidth}
    \centering
    \includegraphics[width=\linewidth]{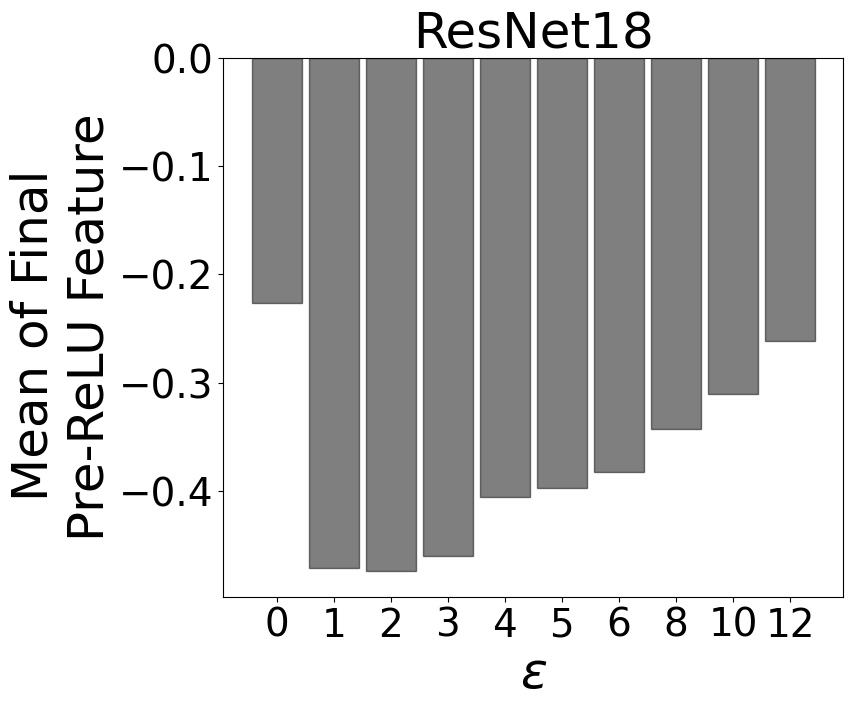}
\end{subfigure}
\begin{subfigure}{.45\linewidth}
    \centering
    \includegraphics[width=\linewidth]{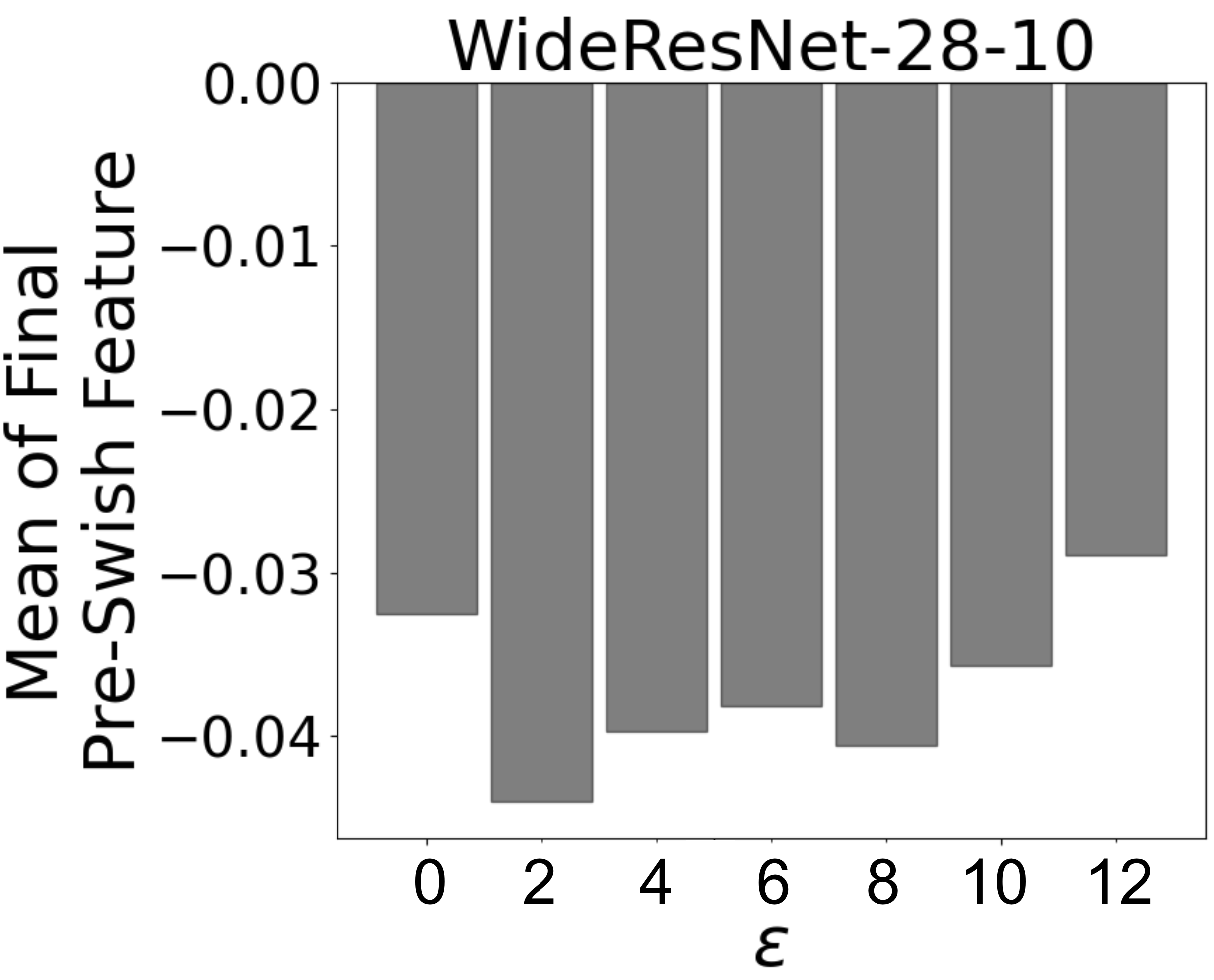}
\end{subfigure}
\caption{Mean of Final Pre-ReLU / Pre-Swish features for increasingly robust ResNet18 \emph{(left)} and WideResNet-28-10 \emph{(right)} models.}
\label{fig:mini_relu_mean}
\vspace{-0.5em}
\end{figure}


\textbf{Discussions: }We observe in Figure \ref{fig:hist}, the first two curves which denote the average $L_{\infty}$ Lipschitz constant of filters in the first two layers of ResNet18, take much larger values when compared to the curves corresponding to the rest of the layers. The large separation between the first two layers and remaining layers is again observed for WideResNet-28-10.
The plot of maximum $L_{\infty}$ Lipschitz constant of filters of each layer in Appendix \ref{app:change} also displays a similar separation between the first two layers and other layers. Moreover, it depicts that the curves of the first and second layer monotonically increase for ResNet18 but not for WideResNet-28-10. From the above, we infer the following: (1) the first two layers are primarily responsible for propagating perturbations from input to output, and (2) wider networks may realize greater adversarial robustness from implicit regularization of their first two layers. Further, our conclusion on the  filter's $L_{\infty}$ Lipschitz constants of the first two layers playing an outsized role, agrees with our previous observations in Section \ref{sec:quant}. Where, quantization of intermediate features after these layers provided the greatest increases in accuracy. This leads us to the following conjecture: the first two layers play an important role in robustness compared to the other layers. When designing robust techniques, special attention needs to be devoted to these layers in order to have a far reaching impact.

\section{Training with Larger Perturbations and Common-Corruptions} \label{sec:ood_all}
\textbf{Motivation: }While training with a spectrum of $\epsilon$ values can uncover new techniques, otherwise not apparent at a single $\epsilon$, it can have potential downsides. An alternate way to assess the robustness of a DNN model is to evaluate on common corruptions like image noise. When trained in an adversarial setting  we expect the following - AT, albeit unintentionally, should increase the general robustness of the model against corruptions like Gaussian noise, weather shifts and alike as discussed in \citet{hendrycks2019benchmarking}. Even though AT is a standard comparison benchmark when it comes to techniques for increasing robustness in such common corruptions setting, the further analysis into testing the limits of this is often overlooked. The only work that tests the limits \citep{kireev2021effectiveness} has previously reported that $1$-robust models perform better than $8$-robust models.

\textbf{Experiments: }To this end, we test increasingly robust models on images with common corruptions from the CIFAR-10-C dataset \citep{hendrycks2019benchmarking}. The complete results on each corruption is in Appendix \ref{app:OOD_all}. We present the average accuracy across the full dataset in Table \ref{tab:mini_corruptions}. We notice another facet of perturbation strength - models trained with smaller $\epsilon$'s obtain better robustness across all kinds of corruptions -- which is not what one would expect. 
Further, the accuracy on corrupted images uniformly decreases after $\epsilon=1$ or $2$ on almost all corruptions. On average, the $2$-robust models have the highest robustness (for both ResNet18 and WideResNet-28-10), followed by the $1$-robust model for ResNet18 and the $4$-robust model for WideResNet-28-10. 

\textbf{Discussions: }We understand this peculiar behaviour in terms of the features before ReLU functions and in particular the final Pre-ReLU/Pre-Swish features. Unlike adversarial attacked images which have small bounded perturbations, corrupted images,  have numerically larger perturbations. These perturbations when propagated through the model are more likely to flip an input to a ReLU neuron from negative to positive (or vice versa) when it is closer to zero. Consequently, we notice that the mean of the final pre-ReLU features (for ResNet18) and pre-Swish features (for WideResNet-28-10) on the complete CIFAR-10-C dataset are further away from zero for lower strength models (see Figure \ref{fig:mini_relu_mean}). 
The highly biased positions of these final pre-activation features is likely to be responsible for the increased robustness of the lower $\epsilon$-robust models. Moreover, this reveals that it may not always be helpful to train with increasing $\epsilon$, particularly from the vantage point of images corrupted with common noise.

\section{Related Work}
AT has been discussed in a vast body of work. Here, we only describe literature relevant to our findings.


\textbf{On varying perturbation strength in AT: }On the attack side, recent papers have explored robustness to unseen attacks \citep{stutz2020confidence, laidlaw2020perceptual}. Our proposed quantization is complementary to these methods (and standard AT and its variants) and unlike these methods does not explicitly need to be calibrated to unseen attacks. 

For defenses, while few papers \citep{balaji2019instance, cheng2020cat, ding2018mma} have proposed AT algorithms that utilize varying perturbation strengths for each sample, they are still usually constrained by an upper bound on perturbation strength. Thus, these works are still restricted to feedback from a model when trained within that upper bound. As far as we are aware, there is a dearth of work that employs feedback from a spectrum of perturbation strengths or upper bounds to improve AT. In future work, we plan to execute the aforementioned perturbation-customized algorithms at various upper bounds to gain further insight into improvements for AT.


\textbf{On quantization: }Quantization and transformations of input images have been explored in \citet{buckman2018thermometer, guo2017countering} but are only able to provide robustness gains because of obfuscated gradients, and hence are unable to stand against an adaptive BPDA attack \citep{athalye2018obfuscated}. Further, recent work on intermediate feature transformations based on Neural ODEs \citep{kang2021stable, chen2021towards} have been shown to also not provide any additional benefits against adaptive attacks \citep{croce2022evaluating}. In contrast, by simply applying quantization, we improve robustness against both BPDA and Transfer PGD.


\textbf{On convolution filters and AT: }Concurrent work \citep{gavrikov2022adversarial} has analyzed robust models from the viewpoint of the models' convolution filters. They analyze the sparsity and diversity of filters to conclude that the first layer plays a very important role. While we concur with their analysis of the importance of the first layer, we identify exactly how the first layer contributes to perturbation propagation by analyzing filter norms across a spectrum of pertubation strengths. Further, we identify, again across various perturbations strengths, that the second layer appears just as important through both the filter norms and the effect of quantization after both layers.

\section{Conclusions and Future Work} 
In this paper, we analyzed AT models trained at a spectrum of perturbation strengths and identified new techniques (overdesigning, quantization, sensitivity of convolutional filter weights) not apparent from standard analysis at just a single defense perturbation strength. We explored the effects of these techniques on adversarial robustness of ResNet and WideResNet models to seen and unseen, standard and adaptive attacks on CIFAR-10. Finally, we discovered a potential downside to training at various perturbation strengths, in the form of a drop in robustness to common corruptions. In future work, we aim to study perturbation-customized AT variants and test-time defenses at a spectrum of perturbations, and identify methods to regularize the first two layers to further improve robustness in AT models.

\bibliography{intriguing_references}
\bibliographystyle{icml2022}

\newpage
\appendix
\onecolumn
\newpage
\section*{Appendix: Experimental Details and Additional Results}

Our code and models can be found at the following link: \url{https://github.com/perturb-spectrum/analysis_and_methods}.

\section{Adversarially Trained Models' Performance and Hyperparameters} \label{app:individual}
\subsection{Adversarially Trained Models}
\begin{table}[H]
\tiny
\centering
\begin{tabular}{c|c|c|c|c|c|c|c|c|c|c|c|c|c}
Accuracy (\%) & \multicolumn{7}{c}{$\epsilon$} \\
\cline{2-14}
 & $0$ & $1$ & $2$ & $3$ & $4$ & $5$ & $6$ & $7$ & $8$ & $9$ & $10$ & $11$ & $12$ \\
\hline 

clean & 85.58 & 88.25 & 88.79 & 88.16 & 88.09 & 87.95 & 86.35 & 84.66 & 84.63 & 83.23 & 83.32 & 83.33 & 83.5  \\
PGD$_{\epsilon}$-20 & 85.58 & 84.57 & 77.83 & 71.32 & 65.51 & 60.54 & 55.71 & 52.06 & 48.75 & 45.56 & 43.16 & 41.69 & 39.06  \\ \hline

\end{tabular}
\caption{Clean and adversarial performance of ResNet-18 models on increasing values of $\epsilon$.}
\end{table}

\begin{table}[H]
\tiny
\centering
\begin{tabular}{c|c|c|c|c|c|c|c}
Accuracy (\%) & \multicolumn{7}{c}{$\epsilon$} \\
\cline{2-8}
 & $0$ & $2$ & $4$ & $6$ & $8$ & $10$ & $12$ \\
\hline

clean & 85.45 & 88.41 & 88.13 & 88.66 & 85.23 & 83.51 & 82.83  \\
PGD$_{\epsilon}$-20 & 85.45 & 81.77 & 70.84 & 62.03 & 53.75 & 47.54 & 42.17  \\ \hline

\end{tabular}
\caption{Clean and adversarial performance of WideResNet-28-10 models on increasing values of $\epsilon$.
}
\end{table}

\begin{table}[H]
\tiny
\centering
\begin{tabular}{c|c|c|c|c}
Hyperparameter & \multicolumn{2}{c|}{ResNet18} & \multicolumn{2}{c}{WideResNet-28-10} \\
\cline{2-5}
& Standard Training & AT & Standard Training & AT \\
\hline 

batch size & 512 & 256 & 512 & 128 \\ \hline
optimizer & \multicolumn{4}{c}{SGD with momentum=$0.9$ and weight decay=$2\times10^{-4}$} \\ \hline
learning rate schedule & \multicolumn{4}{c}{starting at $0.1$ and divided by $10$ at epochs 75, 90, and 100.}  \\ \hline
PGD Step Size & - & $\frac{\epsilon}{8} \times \frac{2}{255}$ & - & $\frac{\epsilon}{8} \times \frac{2}{255}$ \\ \hline
PGD Random Start & - & True & - & True \\ \hline
total epochs & 50 & 200 & 50 & 200 \\ \hline
epoch of best checkpoint & 19 & 76 & 18 & 80 \\ \hline

\end{tabular}
\caption{Hyperparameters for Standard and AT with ResNet18 and WideResNet-28-10.}
\label{BPDA}
\end{table}






\section{Overdesigning for Robust Generalization}\label{app:gen}
\begin{figure}[H]
\centering
\begin{subfigure}{.33\textwidth}
    \centering
    \includegraphics[width=\linewidth]{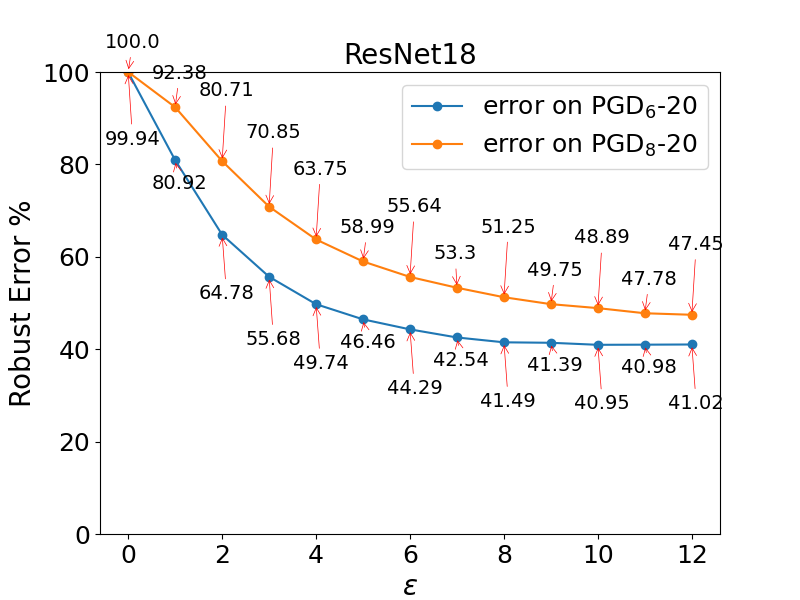}
\end{subfigure}
\begin{subfigure}{.33\textwidth}
    \centering
    \includegraphics[width=\linewidth]{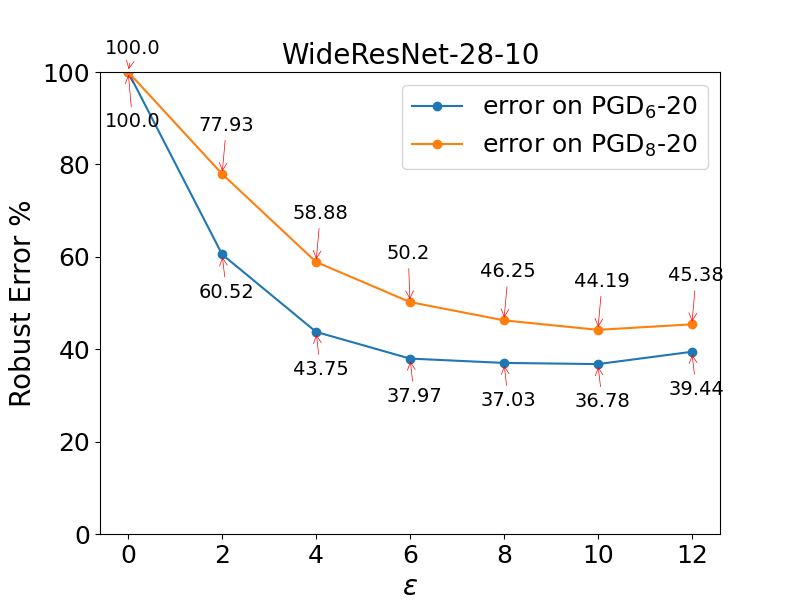}
\end{subfigure}
\caption{Robust error (= error on PGD$_{\delta}$-20 on test set) of increasingly $\epsilon$-robust ResNet18 \emph{(left)} and WideResNet-28-10 \emph{(right)} models for $\delta \in \{6, 8\}$.}
\label{fig:g_err_appendix}
\end{figure}

Figure \ref{fig:g_err_appendix} depicts the robust error for increasingly $\epsilon$-robust ResNet18 and WideResNet-28 at attack perturbation strengths $\delta=6$ and $8$. The trend is similar to Figure \ref{fig:g_err_mini}. Specifically, we notice that the $9$-robust model generalizes best to PGD$_{6}$-20 for ResNet18; the $10$-robust model generalizes best to PGD$_{6}$-20 for WideResNet-28-10 continuing the trend in $\epsilon_{\text{best}}-\delta$ discussed in Section \ref{sec:gen}.

\section{Intermediate feature Quantization}\label{app:quant}

\subsection{BPDA}
We table the complete set of accuracies of $\epsilon$ robust ResNet18 and WideResNet-28-10 models on BPDA$_{\delta}$, for various $\epsilon$ and $\delta$, in Tables \ref{tab:BPDA_appendix} and \ref{tab:BPDA_wide_appendix} respectively. 
\begin{table}[H]
\tiny
\centering
\begin{tabular}{c|c|c|c|c|c|c|c|c|c|c|c|c|c|c}
$\delta$ & Quant. & \multicolumn{7}{c}{$\epsilon$} \\
\cline{3-15}
 & after & $0$ & $1$ & $2$ & $3$ & $4$ & $5$ & $6$ & $7$ & $8$ & $9$ & $10$ & $11$ & $12$ \\
 & layer: & & & & & & & & & & & & & \\
\hline

2 & none & \cellcolor{gray!20}18.15 & \cellcolor{gray!20}71.84 & \cellcolor{gray!20}77.85 & \cellcolor{gray!20}79.01 & \cellcolor{gray!20}79.08 & \cellcolor{gray!20}78.75 & \cellcolor{gray!20}77.77 & \cellcolor{gray!20}77.05 & \cellcolor{gray!20}76.69 & \cellcolor{gray!20}75.57 & \cellcolor{gray!20}74.13 & \cellcolor{gray!20}72.28 & \cellcolor{gray!20}71.01  \\
 & conv0 & \cellcolor{blue!40}21.04& \cellcolor{red!6}71.53& \cellcolor{red!6}77.29& \cellcolor{red!6}77.92& \cellcolor{red!6}78.29& \cellcolor{red!6}77.96& \cellcolor{red!6}76.15& \cellcolor{red!6}75.36& \cellcolor{red!6}75.25& \cellcolor{red!6}73.98& \cellcolor{red!6}71.28& \cellcolor{red!6}69.54& 68.41  \\
 & layer1 & \cellcolor{red!6}17.57& \cellcolor{red!6}71.78& \cellcolor{blue!40}78.00& \cellcolor{blue!40}79.05& \cellcolor{red!6}79.07& \cellcolor{blue!7}78.77& \cellcolor{red!6}77.67& \cellcolor{red!6}76.66& \cellcolor{red!6}76.38& \cellcolor{red!6}74.62& \cellcolor{red!6}73.61& \cellcolor{red!6}71.78& 70.28  \\
 & layer2 & \cellcolor{blue!7}18.21& \cellcolor{blue!8}71.90& 77.85 & \cellcolor{red!6}78.84& \cellcolor{blue!8}79.15& \cellcolor{red!6}78.64& \cellcolor{blue!16}77.81& \cellcolor{red!6}76.96& \cellcolor{red!6}76.53& \cellcolor{blue!7}75.59& \cellcolor{red!6}73.44& \cellcolor{red!6}71.74& 70.14  \\
 & layer3 & \cellcolor{red!6}18.04& \cellcolor{blue!40}72.11& \cellcolor{blue!7}77.86& \cellcolor{red!6}78.88& \cellcolor{blue!40}79.42& \cellcolor{blue!40}78.86& \cellcolor{blue!40}77.87& \cellcolor{blue!40}77.14& \cellcolor{blue!7}76.71& \cellcolor{blue!40}75.81& \cellcolor{blue!40}74.17& \cellcolor{blue!40}72.49& 70.95  \\
 & layer4 & \cellcolor{blue!7}18.20& \cellcolor{blue!29}72.04& \cellcolor{blue!21}77.93& \cellcolor{red!6}79.00& \cellcolor{blue!14}79.20& \cellcolor{red!6}78.58& \cellcolor{red!6}77.67& 77.05 & \cellcolor{blue!40}76.91& \cellcolor{blue!11}75.64& \cellcolor{red!6}74.11& \cellcolor{blue!9}72.33& 70.79  \\
\hline
4 & none & \cellcolor{gray!20}1.49 & \cellcolor{gray!20}41.50 & \cellcolor{gray!20}56.39 & \cellcolor{gray!20}62.54 & \cellcolor{gray!20}65.46 & \cellcolor{gray!20}67.42 & \cellcolor{gray!20}67.82 & \cellcolor{gray!20}68.16 & \cellcolor{gray!20}68.50 & \cellcolor{gray!20}67.61 & \cellcolor{gray!20}66.88 & \cellcolor{gray!20}66.08 & \cellcolor{gray!20}65.20  \\
 & conv0 & \cellcolor{blue!40}2.26& \cellcolor{blue!40}44.01& \cellcolor{blue!40}57.47& \cellcolor{blue!40}62.95& \cellcolor{red!6}65.27& \cellcolor{red!6}66.65& \cellcolor{red!6}66.54& \cellcolor{red!6}66.80& \cellcolor{red!6}67.18& \cellcolor{red!6}66.24& \cellcolor{red!6}64.32& \cellcolor{red!6}63.37& \cellcolor{red!6}62.92 \\
 & layer1 & \cellcolor{red!6}1.35& \cellcolor{blue!7}41.72& \cellcolor{blue!18}56.88& \cellcolor{blue!35}62.90& \cellcolor{blue!26}65.72& \cellcolor{red!6}67.40& \cellcolor{blue!13}67.89& \cellcolor{red!6}68.06& \cellcolor{red!6}68.12& \cellcolor{red!6}67.10& \cellcolor{red!6}66.63& \cellcolor{red!6}65.50& \cellcolor{red!6}64.87 \\
 & layer2 & \cellcolor{red!6}1.47& \cellcolor{blue!7}41.82& \cellcolor{blue!7}56.53& \cellcolor{blue!15}62.70& \cellcolor{blue!40}65.85& \cellcolor{blue!40}67.68& \cellcolor{blue!40}68.03& \cellcolor{red!6}68.02& \cellcolor{red!6}68.23& \cellcolor{red!6}67.60& \cellcolor{red!6}66.65& \cellcolor{red!6}65.72& \cellcolor{red!6}64.85 \\
 & layer3 & \cellcolor{red!6}1.45& \cellcolor{blue!7}41.54& \cellcolor{blue!7}56.59& \cellcolor{blue!13}62.68& \cellcolor{blue!36}65.82& \cellcolor{blue!38}67.67& \cellcolor{blue!38}68.02& \cellcolor{blue!40}68.61& \cellcolor{blue!40}68.54& \cellcolor{blue!40}67.97& \cellcolor{blue!40}67.28& \cellcolor{blue!13}66.15& \cellcolor{blue!31}65.55 \\
 & layer4 & \cellcolor{red!6}1.47& \cellcolor{blue!7}41.73& \cellcolor{blue!7}56.49& \cellcolor{blue!15}62.70& \cellcolor{blue!15}65.61& \cellcolor{blue!9}67.48& \cellcolor{blue!15}67.90& \cellcolor{blue!7}68.21& \cellcolor{blue!10}68.51& \cellcolor{blue!22}67.81& \cellcolor{blue!20}67.08& \cellcolor{blue!40}66.29& \cellcolor{blue!40}65.65 \\
\hline
6 & none & \cellcolor{gray!20}0.05 & \cellcolor{gray!20}19.10 & \cellcolor{gray!20}35.31 & \cellcolor{gray!20}44.50 & \cellcolor{gray!20}50.31 & \cellcolor{gray!20}53.50 & \cellcolor{gray!20}55.59 & \cellcolor{gray!20}57.45 & \cellcolor{gray!20}58.60 & \cellcolor{gray!20}58.66 & \cellcolor{gray!20}59.12 & \cellcolor{gray!20}59.03 & \cellcolor{gray!20}58.97  \\
 & conv0 & \cellcolor{blue!40}0.16& \cellcolor{blue!40}21.23& \cellcolor{blue!40}36.85& \cellcolor{blue!40}45.87& \cellcolor{blue!40}51.45& \cellcolor{blue!40}54.26& \cellcolor{blue!15}55.77& \cellcolor{red!6}57.02& \cellcolor{red!6}58.22& \cellcolor{red!6}57.93& \cellcolor{red!6}57.11& \cellcolor{red!6}57.18& \cellcolor{red!6}57.49 \\
 & layer1 & \cellcolor{blue!10}0.08& \cellcolor{red!6}19.04& \cellcolor{blue!7}35.43& \cellcolor{blue!20}45.20& \cellcolor{blue!21}50.91& \cellcolor{blue!32}54.12& \cellcolor{blue!36}56.01& \cellcolor{blue!18}57.63& \cellcolor{blue!9}58.70& \cellcolor{red!6}58.60& \cellcolor{red!6}58.90& \cellcolor{red!6}58.97& \cellcolor{red!6}58.83 \\
 & layer2 & \cellcolor{red!6}0.04& \cellcolor{red!6}19.08& \cellcolor{blue!7}35.61& \cellcolor{blue!7}44.74& \cellcolor{blue!17}50.80& \cellcolor{blue!18}53.85& \cellcolor{blue!27}55.91& \cellcolor{blue!22}57.67& \cellcolor{red!6}58.59& \cellcolor{blue!32}59.04& \cellcolor{red!6}59.05& \cellcolor{blue!10}59.21& \cellcolor{blue!11}59.18 \\
 & layer3 & 0.05 & \cellcolor{blue!7}19.18& \cellcolor{red!6}35.19& \cellcolor{blue!11}44.90& \cellcolor{blue!12}50.68& \cellcolor{blue!23}53.94& \cellcolor{blue!40}56.05& \cellcolor{blue!40}57.84& \cellcolor{blue!27}58.90& \cellcolor{blue!40}59.13& \cellcolor{blue!29}59.47& \cellcolor{blue!18}59.33& \cellcolor{blue!40}59.68 \\
 & layer4 & 0.05 & \cellcolor{blue!7}19.31& \cellcolor{blue!7}35.32& \cellcolor{red!6}44.45& \cellcolor{blue!7}50.41& \cellcolor{blue!15}53.80& \cellcolor{blue!36}56.01& \cellcolor{blue!22}57.67& \cellcolor{blue!40}59.04& \cellcolor{blue!34}59.06& \cellcolor{blue!40}59.60& \cellcolor{blue!40}59.69& \cellcolor{blue!38}59.65 \\
\hline
8 & none & \cellcolor{gray!20}0.00 & \cellcolor{gray!20}7.74 & \cellcolor{gray!20}19.37 & \cellcolor{gray!20}29.25 & \cellcolor{gray!20}36.20 & \cellcolor{gray!20}41.00 & \cellcolor{gray!20}44.29 & \cellcolor{gray!20}46.76 & \cellcolor{gray!20}48.77 & \cellcolor{gray!20}50.13 & \cellcolor{gray!20}51.08 & \cellcolor{gray!20}52.21 & \cellcolor{gray!20}52.53  \\
 & conv0 & 0.00 & \cellcolor{blue!40}8.72& \cellcolor{blue!40}20.78& \cellcolor{blue!40}30.72& \cellcolor{blue!40}37.40& \cellcolor{blue!40}41.65& \cellcolor{blue!7}44.37& \cellcolor{blue!7}46.77& \cellcolor{red!6}48.74& \cellcolor{red!6}50.05& \cellcolor{red!6}50.01& \cellcolor{red!6}50.57& \cellcolor{red!6}51.15 \\
 & layer1 & 0.00 & \cellcolor{red!6}7.51& \cellcolor{blue!7}19.38& \cellcolor{blue!14}29.79& \cellcolor{blue!20}36.82& \cellcolor{blue!37}41.61& \cellcolor{blue!29}44.78& \cellcolor{blue!12}47.02& \cellcolor{blue!25}49.28& \cellcolor{blue!11}50.32& \cellcolor{blue!7}51.15& \cellcolor{blue!8}52.44& \cellcolor{red!6}52.41 \\
 & layer2 & 0.00 & \cellcolor{red!6}7.55& \cellcolor{blue!7}19.61& \cellcolor{red!6}29.22& \cellcolor{blue!13}36.60& \cellcolor{blue!36}41.59& \cellcolor{blue!40}44.95& \cellcolor{blue!30}47.40& \cellcolor{blue!27}49.33& \cellcolor{blue!31}50.64& \cellcolor{blue!28}51.54& \cellcolor{blue!10}52.50& \cellcolor{blue!16}52.90 \\
 & layer3 & 0.00 & \cellcolor{red!6}7.28& \cellcolor{blue!7}19.38& \cellcolor{blue!7}29.53& \cellcolor{blue!7}36.32& \cellcolor{blue!31}41.51& \cellcolor{blue!29}44.78& \cellcolor{blue!40}47.60& \cellcolor{blue!40}49.58& \cellcolor{blue!24}50.53& \cellcolor{blue!33}51.63& \cellcolor{blue!15}52.63& \cellcolor{blue!23}53.06 \\
 & layer4 & 0.00 & \cellcolor{red!6}7.62& \cellcolor{red!6}19.30& \cellcolor{red!6}29.19& \cellcolor{blue!7}36.29& \cellcolor{blue!20}41.34& \cellcolor{blue!24}44.70& \cellcolor{blue!24}47.28& \cellcolor{blue!39}49.56& \cellcolor{blue!40}50.78& \cellcolor{blue!40}51.73& \cellcolor{blue!40}53.27& \cellcolor{blue!40}53.42 \\
\hline
10 & none & \cellcolor{gray!20}0.00 & \cellcolor{gray!20}2.66 & \cellcolor{gray!20}9.63 & \cellcolor{gray!20}17.81 & \cellcolor{gray!20}24.41 & \cellcolor{gray!20}29.79 & \cellcolor{gray!20}33.58 & \cellcolor{gray!20}36.81 & \cellcolor{gray!20}39.64 & \cellcolor{gray!20}41.57 & \cellcolor{gray!20}43.21 & \cellcolor{gray!20}45.14 & \cellcolor{gray!20}45.67  \\
 & conv0 & 0.00 & \cellcolor{blue!40}3.12& \cellcolor{blue!40}10.67& \cellcolor{blue!40}18.83& \cellcolor{blue!40}25.17& \cellcolor{blue!30}30.23& \cellcolor{blue!33}34.15& \cellcolor{blue!11}37.08& \cellcolor{red!6}39.29& \cellcolor{blue!7}41.71& \cellcolor{red!6}42.57& \cellcolor{red!6}43.69& \cellcolor{red!6}44.87 \\
 & layer1 & 0.00 & \cellcolor{red!6}2.53& \cellcolor{blue!7}9.68& \cellcolor{blue!15}18.20& \cellcolor{blue!31}25.01& \cellcolor{blue!37}30.33& \cellcolor{blue!40}34.26& \cellcolor{blue!15}37.16& \cellcolor{blue!26}40.21& \cellcolor{blue!10}41.79& \cellcolor{blue!7}43.35& \cellcolor{blue!7}45.33& \cellcolor{blue!7}45.73 \\
 & layer2 & 0.00 & \cellcolor{red!6}2.58& \cellcolor{blue!7}9.75& \cellcolor{blue!7}17.98& \cellcolor{blue!20}24.80& \cellcolor{blue!22}30.11& \cellcolor{blue!30}34.10& \cellcolor{blue!40}37.73& \cellcolor{blue!29}40.26& \cellcolor{blue!40}42.39& \cellcolor{blue!36}43.96& \cellcolor{blue!29}45.93& \cellcolor{blue!20}46.34 \\
 & layer3 & 0.00 & \cellcolor{red!6}2.59& \cellcolor{red!6}9.25& \cellcolor{blue!7}17.91& \cellcolor{blue!7}24.53& \cellcolor{blue!40}30.37& \cellcolor{blue!36}34.20& \cellcolor{blue!32}37.56& \cellcolor{blue!30}40.28& \cellcolor{blue!24}42.08& \cellcolor{blue!31}43.87& \cellcolor{blue!15}45.56& \cellcolor{blue!13}46.10 \\
 & layer4 & 0.00 & \cellcolor{blue!7}2.68& \cellcolor{red!6}9.52& \cellcolor{blue!7}17.99& \cellcolor{blue!7}24.55& \cellcolor{red!6}29.76& \cellcolor{blue!27}34.05& \cellcolor{blue!15}37.17& \cellcolor{blue!40}40.49& \cellcolor{blue!37}42.34& \cellcolor{blue!40}44.04& \cellcolor{blue!40}46.22& \cellcolor{blue!40}46.96 \\
\hline
12 & none & \cellcolor{gray!20}0.00 & \cellcolor{gray!20}0.94 & \cellcolor{gray!20}4.28 & \cellcolor{gray!20}9.97 & \cellcolor{gray!20}15.48 & \cellcolor{gray!20}20.20 & \cellcolor{gray!20}24.16 & \cellcolor{gray!20}27.80 & \cellcolor{gray!20}31.17 & \cellcolor{gray!20}33.49 & \cellcolor{gray!20}35.74 & \cellcolor{gray!20}38.05 & \cellcolor{gray!20}39.27  \\
 & conv0 & 0.00 & \cellcolor{blue!40}1.20& \cellcolor{blue!40}5.28& \cellcolor{blue!40}10.97& \cellcolor{blue!40}16.10& \cellcolor{blue!40}20.99& \cellcolor{blue!21}24.61& \cellcolor{red!6}27.51& \cellcolor{red!6}30.85& \cellcolor{blue!7}33.61& \cellcolor{red!6}35.25& \cellcolor{red!6}36.74& \cellcolor{red!6}38.76 \\
 & layer1 & 0.00 & \cellcolor{red!6}0.82& \cellcolor{blue!7}4.29& \cellcolor{blue!17}10.42& \cellcolor{blue!7}15.60& \cellcolor{blue!29}20.78& \cellcolor{blue!31}24.84& \cellcolor{blue!26}28.24& \cellcolor{blue!19}31.56& \cellcolor{blue!34}34.31& \cellcolor{blue!13}36.11& \cellcolor{blue!24}38.81& \cellcolor{blue!10}39.57 \\
 & layer2 & 0.00 & \cellcolor{red!6}0.86& \cellcolor{blue!7}4.37& \cellcolor{blue!7}10.08& \cellcolor{blue!19}15.79& \cellcolor{blue!21}20.63& \cellcolor{blue!29}24.78& \cellcolor{blue!40}28.47& \cellcolor{blue!40}31.95& \cellcolor{blue!37}34.38& \cellcolor{blue!40}36.81& \cellcolor{blue!38}39.22& \cellcolor{blue!40}40.43 \\
 & layer3 & 0.00 & \cellcolor{red!6}0.83& \cellcolor{blue!7}4.29& \cellcolor{red!6}9.94& \cellcolor{blue!7}15.52& \cellcolor{blue!9}20.38& \cellcolor{blue!40}25.01& \cellcolor{blue!31}28.33& \cellcolor{blue!37}31.91& \cellcolor{blue!32}34.26& \cellcolor{blue!28}36.49& \cellcolor{blue!16}38.56& \cellcolor{blue!13}39.66 \\
 & layer4 & 0.00 & \cellcolor{red!6}0.91& \cellcolor{blue!7}4.34& 9.97 & \cellcolor{blue!9}15.63& \cellcolor{blue!7}20.26& \cellcolor{blue!21}24.62& \cellcolor{blue!27}28.26& \cellcolor{blue!30}31.76& \cellcolor{blue!40}34.43& \cellcolor{blue!34}36.67& \cellcolor{blue!40}39.27& \cellcolor{blue!37}40.36 \\
\hline

\end{tabular}
\caption{Evaluation of quantization after four intermediate features of $\epsilon$-robust \underline{ResNet-18} models on the BPDA$_{\delta}$ attack for various $\epsilon$ and $\delta$ at $\beta=8.0$.}
\label{tab:BPDA_appendix}
\end{table}


\begin{table}[H]
\tiny
\centering
\begin{tabular}{c|c|c|c|c|c|c|c|c}
$\delta$ & Quant. & \multicolumn{7}{c}{$\epsilon$} \\
\cline{3-9}
 & after & $0$ & $2$ & $4$ & $6$ & $8$ & $10$ & $12$ \\
 & layer: & & & & & & \\
\hline

2 & none & \cellcolor{gray!20}2.43 & \cellcolor{gray!20}81.77 & \cellcolor{gray!20}82.67 & \cellcolor{gray!20}81.58 & \cellcolor{gray!20}79.19 & \cellcolor{gray!20}76.23 & \cellcolor{gray!20}71.20  \\
 & init conv & \cellcolor{blue!40}2.48& \cellcolor{red!6}81.62& \cellcolor{blue!32}82.76& \cellcolor{blue!40}81.62& 79.13 & 75.98 & 70.90  \\
 & layer[0] & \cellcolor{red!6}2.11& \cellcolor{blue!40}81.85& \cellcolor{red!6}82.62& \cellcolor{red!6}81.47& 78.93 & 76.11 & 70.98  \\
 & layer[1] & \cellcolor{red!6}2.38& \cellcolor{red!6}81.74& \cellcolor{blue!25}82.74& \cellcolor{red!6}81.57& 79.02 & 76.06 & 70.50  \\
 & layer[2] & \cellcolor{red!6}2.41& \cellcolor{red!6}81.74& \cellcolor{blue!40}82.78& \cellcolor{red!6}81.49& 79.11 & 76.06 & 70.90  \\
\hline
4 & none & \cellcolor{gray!20}0.00 & \cellcolor{gray!20}62.12 & \cellcolor{gray!20}70.87 & \cellcolor{gray!20}72.80 & \cellcolor{gray!20}71.81 & \cellcolor{gray!20}69.95 & \cellcolor{gray!20}66.20  \\
 & init conv & 0.00 & \cellcolor{red!6}61.73& \cellcolor{red!6}70.80& \cellcolor{blue!13}72.91& \cellcolor{red!6}71.53& \cellcolor{red!6}69.62& 65.81  \\
 & layer[0] & 0.00 & \cellcolor{blue!40}62.39& \cellcolor{blue!18}70.95& \cellcolor{blue!40}73.12& \cellcolor{blue!7}71.82& \cellcolor{red!6}69.74& 66.06  \\
 & layer[1] & 0.00 & \cellcolor{blue!7}62.14& \cellcolor{blue!7}70.90& \cellcolor{blue!9}72.88& \cellcolor{blue!7}71.82& \cellcolor{red!6}69.80& 65.82  \\
 & layer[2] & 0.00 & \cellcolor{blue!7}62.17& \cellcolor{blue!40}71.04& \cellcolor{blue!7}72.82& \cellcolor{blue!40}71.90& \cellcolor{blue!40}70.05& 65.88  \\
\hline
6 & none & \cellcolor{gray!20}0.00 & \cellcolor{gray!20}39.50 & \cellcolor{gray!20}56.25 & \cellcolor{gray!20}62.01 & \cellcolor{gray!20}62.91 & \cellcolor{gray!20}63.28 & \cellcolor{gray!20}60.54  \\
 & init conv & 0.00 & \cellcolor{red!6}39.30& \cellcolor{red!6}56.19& \cellcolor{blue!15}62.15& \cellcolor{blue!7}62.95& \cellcolor{red!6}62.90& \cellcolor{blue!32}60.72 \\
 & layer[0] & 0.00 & \cellcolor{blue!40}40.28& \cellcolor{blue!40}56.61& \cellcolor{blue!12}62.13& \cellcolor{blue!12}63.01& \cellcolor{red!6}63.03& \cellcolor{blue!9}60.59 \\
 & layer[1] & 0.00 & \cellcolor{blue!11}39.73& \cellcolor{blue!8}56.33& \cellcolor{blue!35}62.34& \cellcolor{blue!8}62.98& \cellcolor{blue!9}63.33& \cellcolor{red!6}60.44 \\
 & layer[2] & 0.00 & \cellcolor{blue!7}39.61& \cellcolor{blue!7}56.32& \cellcolor{blue!40}62.38& \cellcolor{blue!40}63.23& \cellcolor{blue!40}63.50& \cellcolor{blue!40}60.76 \\
\hline
8 & none & \cellcolor{gray!20}0.00 & \cellcolor{gray!20}22.08 & \cellcolor{gray!20}41.20 & \cellcolor{gray!20}49.83 & \cellcolor{gray!20}53.72 & \cellcolor{gray!20}55.81 & \cellcolor{gray!20}54.67  \\
 & init conv & 0.00 & \cellcolor{red!6}21.88& \cellcolor{red!6}41.17& \cellcolor{red!6}49.76& \cellcolor{blue!11}53.84& \cellcolor{red!6}55.51& \cellcolor{red!6}54.46 \\
 & layer[0] & 0.00 & \cellcolor{blue!40}22.56& \cellcolor{blue!40}41.69& \cellcolor{blue!29}49.99& \cellcolor{blue!40}54.13& \cellcolor{blue!16}55.85& \cellcolor{blue!7}54.69 \\
 & layer[1] & 0.00 & \cellcolor{blue!7}22.12& \cellcolor{blue!18}41.43& \cellcolor{blue!40}50.05& \cellcolor{blue!34}54.07& \cellcolor{red!6}55.69& \cellcolor{red!6}54.57 \\
 & layer[2] & 0.00 & \cellcolor{blue!7}22.14& \cellcolor{blue!10}41.33& \cellcolor{blue!30}50.00& \cellcolor{blue!36}54.09& \cellcolor{blue!40}55.91& \cellcolor{blue!40}54.99 \\
\hline
10 & none & \cellcolor{gray!20}0.00 & \cellcolor{gray!20}11.15 & \cellcolor{gray!20}28.19 & \cellcolor{gray!20}38.37 & \cellcolor{gray!20}44.65 & \cellcolor{gray!20}47.70 & \cellcolor{gray!20}48.12  \\
 & init conv & 0.00 & \cellcolor{red!6}11.11& \cellcolor{red!6}28.13& \cellcolor{blue!7}38.41& \cellcolor{red!6}44.64& \cellcolor{blue!8}47.80& \cellcolor{blue!13}48.34 \\
 & layer[0] & 0.00 & \cellcolor{blue!40}11.44& \cellcolor{blue!40}28.55& \cellcolor{blue!40}38.89& \cellcolor{blue!40}45.00& \cellcolor{blue!40}48.16& \cellcolor{blue!36}48.70 \\
 & layer[1] & 0.00 & \cellcolor{red!6}11.02& \cellcolor{blue!14}28.32& \cellcolor{blue!39}38.88& \cellcolor{blue!24}44.86& \cellcolor{blue!16}47.89& \cellcolor{blue!40}48.76 \\
 & layer[2] & 0.00 & \cellcolor{red!6}11.09& \cellcolor{blue!20}28.37& \cellcolor{blue!26}38.71& \cellcolor{blue!21}44.84& \cellcolor{blue!36}48.12& \cellcolor{blue!32}48.64 \\
\hline
12 & none & \cellcolor{gray!20}0.00 & \cellcolor{gray!20}5.18 & \cellcolor{gray!20}17.54 & \cellcolor{gray!20}28.08 & \cellcolor{gray!20}35.93 & \cellcolor{gray!20}39.84 & \cellcolor{gray!20}42.09  \\
 & init conv & 0.00 & \cellcolor{red!6}5.15& \cellcolor{blue!7}17.66& \cellcolor{blue!11}28.23& \cellcolor{red!6}35.84& \cellcolor{blue!7}39.90& \cellcolor{blue!7}42.20 \\
 & layer[0] & 0.00 & \cellcolor{blue!40}5.35& \cellcolor{blue!40}18.44& \cellcolor{blue!40}28.62& \cellcolor{blue!27}36.31& \cellcolor{blue!7}39.93& \cellcolor{blue!23}42.42 \\
 & layer[1] & 0.00 & \cellcolor{red!6}5.14& \cellcolor{blue!15}17.90& \cellcolor{blue!17}28.32& \cellcolor{blue!37}36.45& \cellcolor{blue!40}40.54& \cellcolor{blue!35}42.59 \\
 & layer[2] & 0.00 & \cellcolor{red!6}5.03& \cellcolor{blue!7}17.72& \cellcolor{blue!24}28.41& \cellcolor{blue!40}36.49& \cellcolor{blue!31}40.40& \cellcolor{blue!40}42.65 \\
\hline

\end{tabular}
\caption{Evaluation of quantization after four intermediate features of $\epsilon$-robust \underline{WideResNet-28-10} models on the BPDA$_{\delta}$ attack for various $\epsilon$ and $\delta$ at $\beta=8.0$.
}
\label{tab:BPDA_wide_appendix}
\end{table}

\subsection{Transfer PGD with Ablations}
We table the complete set of accuracies of $\epsilon$ robust ResNet18 and WIdeResNet-28-10 models on Transfer PGD$_{\delta}$-20 attack, for various $\epsilon$ and $\delta$, at quantization scaling factor of $\beta=8.0$ in Tables \ref{tab:TPGD_appendix} and \ref{tab:TPGD_wide_appendix} respectively.

We repeat our experiments with ResNet18 models for quantization scaling factors of $\beta=6.0, 8.0, 10.0$ and $12.0$ and display all accuracies for various $\epsilon$ and $\delta$ in Tables \ref{tab:TPGD_appendix_4}, \ref{tab:TPGD_appendix_6}, \ref{tab:TPGD_appendix_10}, and \ref{tab:TPGD_appendix_12} respectively.

We repeat our experiments again with WideResNet-28-10 models for quantization scaling factors of $\beta=6.0, 8.0, 10.0$ and $12.0$ and display all accuracies for various $\epsilon$ and $\delta$ in Tables \ref{tab:TPGD_wide_appendix_4}, \ref{tab:TPGD_wide_appendix_6}, \ref{tab:TPGD_wide_appendix_10}, and \ref{tab:TPGD_wide_appendix_12} respectively.

We continue to observe increase in robustness to seen and unseen BPDA and Transfer PGD in all the below tables.
\begin{table}[H]
\tiny
\centering
\begin{tabular}{c|c|c|c|c|c|c|c|c|c|c|c|c|c|c}
$\delta$ & Quant. & \multicolumn{7}{c}{$\epsilon$} \\
\cline{3-15}
 & after & $0$ & $1$ & $2$ & $3$ & $4$ & $5$ & $6$ & $7$ & $8$ & $9$ & $10$ & $11$ & $12$ \\
 & layer: & & & & & & & & & & & & & \\
\hline

2 & none & \cellcolor{gray!20}18.15 & \cellcolor{gray!20}71.84 & \cellcolor{gray!20}77.85 & \cellcolor{gray!20}79.01 & \cellcolor{gray!20}79.08 & \cellcolor{gray!20}78.75 & \cellcolor{gray!20}77.77 & \cellcolor{gray!20}77.05 & \cellcolor{gray!20}76.69 & \cellcolor{gray!20}75.57 & \cellcolor{gray!20}74.13 & \cellcolor{gray!20}72.28 & \cellcolor{gray!20}71.01  \\
 & conv0 & \cellcolor{blue!40}31.83& \cellcolor{blue!40}75.11& \cellcolor{blue!40}79.03& \cellcolor{blue!25}79.36& \cellcolor{blue!32}79.42& \cellcolor{red!6}78.68& \cellcolor{red!6}77.20& \cellcolor{red!6}76.17& \cellcolor{red!6}75.88& \cellcolor{red!6}74.43& \cellcolor{red!6}71.84& \cellcolor{red!6}70.39& \cellcolor{red!6}68.90 \\
 & layer1 & \cellcolor{blue!7}18.57& \cellcolor{blue!10}72.73& \cellcolor{blue!19}78.43& \cellcolor{blue!40}79.56& \cellcolor{blue!21}79.31& \cellcolor{blue!40}78.98& \cellcolor{blue!21}77.88& \cellcolor{red!6}76.83& \cellcolor{red!6}76.46& \cellcolor{red!6}74.78& \cellcolor{red!6}73.61& \cellcolor{red!6}71.82& \cellcolor{red!6}70.30 \\
 & layer2 & \cellcolor{blue!7}18.60& \cellcolor{blue!7}72.38& \cellcolor{blue!7}78.01& \cellcolor{blue!7}79.09& \cellcolor{blue!21}79.31& \cellcolor{blue!7}78.76& \cellcolor{blue!36}77.95& \cellcolor{red!6}77.00& \cellcolor{blue!26}76.71& \cellcolor{blue!7}75.58& \cellcolor{red!6}73.48& \cellcolor{red!6}71.71& \cellcolor{red!6}70.15 \\
 & layer3 & \cellcolor{blue!7}18.29& \cellcolor{blue!7}72.09& \cellcolor{red!6}77.84& \cellcolor{red!6}78.89& \cellcolor{blue!40}79.50& \cellcolor{blue!29}78.92& \cellcolor{blue!40}77.97& \cellcolor{blue!40}77.12& \cellcolor{blue!40}76.72& \cellcolor{blue!40}75.83& \cellcolor{blue!40}74.23& \cellcolor{blue!40}72.40& \cellcolor{red!6}70.97 \\
\hline

\end{tabular}
\caption{Evaluation of quantization on all four intermediate features of $\epsilon$-robust \underline{ResNet18} models on the Transfer PGD$_{\delta}$-20 attack for various $\epsilon$ and $\delta$ at $\underline{\beta=8.0}$. 
}
\label{tab:TPGD_appendix}
\end{table}
\begin{table}[H]
\tiny
\centering
\begin{tabular}{c|c|c|c|c|c|c|c|c|c|c|c|c|c|c}
$\delta$ & Quant. & \multicolumn{7}{c}{$\epsilon$} \\
\cline{3-15}
 & after & $0$ & $1$ & $2$ & $3$ & $4$ & $5$ & $6$ & $7$ & $8$ & $9$ & $10$ & $11$ & $12$ \\
 & layer: & & & & & & & & & & & & & \\
\hline

2 & none & \cellcolor{gray!20}18.15 & \cellcolor{gray!20}71.84 & \cellcolor{gray!20}77.85 & \cellcolor{gray!20}79.01 & \cellcolor{gray!20}79.08 & \cellcolor{gray!20}78.75 & \cellcolor{gray!20}77.77 & \cellcolor{gray!20}77.05 & \cellcolor{gray!20}76.69 & \cellcolor{gray!20}75.57 & \cellcolor{gray!20}74.13 & \cellcolor{gray!20}72.28 & \cellcolor{gray!20}71.01  \\
 & conv0 & \cellcolor{blue!40}39.35& \cellcolor{blue!36}73.15& \cellcolor{red!6}77.24& \cellcolor{red!6}77.33& \cellcolor{red!6}77.04& \cellcolor{red!6}75.89& \cellcolor{red!6}73.23& \cellcolor{red!6}72.37& \cellcolor{red!6}71.49& \cellcolor{red!6}69.60& \cellcolor{red!6}64.05& \cellcolor{red!6}62.19& \cellcolor{red!6}62.61 \\
 & layer1 & \cellcolor{blue!7}19.00& \cellcolor{blue!40}73.28& \cellcolor{blue!40}78.73& \cellcolor{blue!40}79.63& \cellcolor{blue!7}79.15& \cellcolor{blue!14}78.87& \cellcolor{red!6}77.73& \cellcolor{red!6}76.32& \cellcolor{red!6}75.92& \cellcolor{red!6}73.68& \cellcolor{red!6}73.03& \cellcolor{red!6}70.68& \cellcolor{red!6}69.14 \\
 & layer2 & \cellcolor{blue!7}18.85& \cellcolor{blue!17}72.48& \cellcolor{blue!21}78.33& \cellcolor{blue!19}79.32& \cellcolor{blue!29}79.47& \cellcolor{blue!20}78.92& \cellcolor{blue!7}77.85& \cellcolor{red!6}76.61& \cellcolor{red!6}76.16& \cellcolor{red!6}75.21& \cellcolor{red!6}72.15& \cellcolor{red!6}70.76& \cellcolor{red!6}69.07 \\
 & layer3 & \cellcolor{blue!7}18.25& \cellcolor{blue!7}72.09& \cellcolor{blue!7}77.93& \cellcolor{blue!7}79.05& \cellcolor{blue!40}79.61& \cellcolor{blue!40}79.08& \cellcolor{blue!40}78.27& \cellcolor{blue!40}77.16& \cellcolor{blue!40}76.87& \cellcolor{blue!40}75.77& \cellcolor{blue!40}74.45& \cellcolor{red!6}72.27& \cellcolor{red!6}70.79 \\
\hline

\end{tabular}
\caption{Evaluation of quantization after four intermediate features of $\epsilon$-robust \underline{ResNet-18} models on the Transfer PGD$_{\delta}$-20 attack at $\underline{\beta=4.0}$.}
\label{tab:TPGD_appendix_4}
\end{table}


\begin{table}[H]
\tiny
\centering
\begin{tabular}{c|c|c|c|c|c|c|c|c|c|c|c|c|c|c}
$\delta$ & Quant. & \multicolumn{7}{c}{$\epsilon$} \\
\cline{3-15}
 & after & $0$ & $1$ & $2$ & $3$ & $4$ & $5$ & $6$ & $7$ & $8$ & $9$ & $10$ & $11$ & $12$ \\
 & layer: & & & & & & & & & & & & & \\
\hline

2 & none & \cellcolor{gray!20}18.15 & \cellcolor{gray!20}71.84 & \cellcolor{gray!20}77.85 & \cellcolor{gray!20}79.01 & \cellcolor{gray!20}79.08 & \cellcolor{gray!20}78.75 & \cellcolor{gray!20}77.77 & \cellcolor{gray!20}77.05 & \cellcolor{gray!20}76.69 & \cellcolor{gray!20}75.57 & \cellcolor{gray!20}74.13 & \cellcolor{gray!20}72.28 & \cellcolor{gray!20}71.01  \\
 & conv0 & \cellcolor{blue!40}36.49& \cellcolor{blue!40}75.62& \cellcolor{blue!40}79.02& \cellcolor{blue!23}79.35& \cellcolor{blue!7}79.09& \cellcolor{red!6}78.05& \cellcolor{red!6}76.25& \cellcolor{red!6}75.36& \cellcolor{red!6}74.75& \cellcolor{red!6}73.37& \cellcolor{red!6}70.16& \cellcolor{red!6}68.31& \cellcolor{red!6}67.53 \\
 & layer1 & \cellcolor{blue!7}18.75& \cellcolor{blue!13}73.11& \cellcolor{blue!22}78.51& \cellcolor{blue!40}79.60& \cellcolor{blue!18}79.24& \cellcolor{blue!40}79.01& \cellcolor{blue!23}77.94& \cellcolor{red!6}76.65& \cellcolor{red!6}76.24& \cellcolor{red!6}74.51& \cellcolor{red!6}73.67& \cellcolor{red!6}71.67& \cellcolor{red!6}70.01 \\
 & layer2 & \cellcolor{blue!7}18.70& \cellcolor{blue!7}72.34& \cellcolor{blue!9}78.12& \cellcolor{blue!11}79.18& \cellcolor{blue!36}79.39& \cellcolor{red!6}78.74& \cellcolor{blue!20}77.92& \cellcolor{red!6}76.80& \cellcolor{red!6}76.47& \cellcolor{blue!8}75.62& \cellcolor{red!6}73.11& \cellcolor{red!6}71.51& \cellcolor{red!6}69.74 \\
 & layer3 & \cellcolor{blue!7}18.27& \cellcolor{blue!7}72.09& \cellcolor{blue!7}77.94& \cellcolor{blue!7}79.11& \cellcolor{blue!40}79.42& \cellcolor{blue!15}78.85& \cellcolor{blue!40}78.06& \cellcolor{blue!40}77.19& \cellcolor{blue!40}76.76& \cellcolor{blue!40}75.81& \cellcolor{blue!40}74.15& \cellcolor{blue!40}72.37& \cellcolor{red!6}70.93 \\
\hline

\end{tabular}
\caption{Evaluation of quantization after four intermediate features of $\epsilon$-robust \underline{ResNet-18} models on the Transfer PGD$_{\delta}$-20 attack at $\underline{\beta=6.0}$.}
\label{tab:TPGD_appendix_6}
\end{table}


\begin{table}[H]
\tiny
\centering
\begin{tabular}{c|c|c|c|c|c|c|c|c|c|c|c|c|c|c}
$\delta$ & Quant. & \multicolumn{7}{c}{$\epsilon$} \\
\cline{3-15}
 & after & $0$ & $1$ & $2$ & $3$ & $4$ & $5$ & $6$ & $7$ & $8$ & $9$ & $10$ & $11$ & $12$ \\
 & layer: & & & & & & & & & & & & & \\
\hline

2 & none & \cellcolor{gray!20}18.15 & \cellcolor{gray!20}71.84 & \cellcolor{gray!20}77.85 & \cellcolor{gray!20}79.01 & \cellcolor{gray!20}79.08 & \cellcolor{gray!20}78.75 & \cellcolor{gray!20}77.77 & \cellcolor{gray!20}77.05 & \cellcolor{gray!20}76.69 & \cellcolor{gray!20}75.57 & \cellcolor{gray!20}74.13 & \cellcolor{gray!20}72.28 & \cellcolor{gray!20}71.01  \\
 & conv0 & \cellcolor{blue!40}28.66& \cellcolor{blue!40}74.58& \cellcolor{blue!40}78.99& \cellcolor{blue!40}79.55& \cellcolor{blue!40}79.65& \cellcolor{blue!40}78.98& \cellcolor{red!6}77.49& \cellcolor{red!6}76.61& \cellcolor{red!6}76.26& \cellcolor{red!6}74.94& \cellcolor{red!6}72.81& \cellcolor{red!6}71.17& \cellcolor{red!6}69.93 \\
 & layer1 & \cellcolor{blue!7}18.42& \cellcolor{blue!11}72.62& \cellcolor{blue!13}78.24& \cellcolor{blue!27}79.38& \cellcolor{blue!19}79.36& \cellcolor{blue!29}78.92& \cellcolor{blue!12}77.83& \cellcolor{red!6}76.86& \cellcolor{red!6}76.52& \cellcolor{red!6}75.05& \cellcolor{red!6}73.77& \cellcolor{red!6}71.90& \cellcolor{red!6}70.51 \\
 & layer2 & \cellcolor{blue!7}18.52& \cellcolor{blue!7}72.26& \cellcolor{blue!7}78.02& 79.01 & \cellcolor{blue!9}79.22& \cellcolor{blue!20}78.87& \cellcolor{blue!40}77.96& \cellcolor{red!6}76.90& \cellcolor{blue!40}76.73& \cellcolor{red!6}75.56& \cellcolor{red!6}73.77& \cellcolor{red!6}71.85& \cellcolor{red!6}70.42 \\
 & layer3 & \cellcolor{blue!7}18.27& \cellcolor{blue!7}72.21& \cellcolor{red!6}77.83& \cellcolor{blue!7}79.06& \cellcolor{blue!18}79.34& \cellcolor{blue!13}78.83& \cellcolor{blue!29}77.91& \cellcolor{blue!40}77.22& \cellcolor{red!6}76.61& \cellcolor{blue!40}75.97& \cellcolor{blue!40}74.22& \cellcolor{blue!40}72.47& \cellcolor{blue!40}71.12 \\
\hline

\end{tabular}
\caption{Evaluation of quantization after four intermediate features of $\epsilon$-robust \underline{ResNet-18} models on the Transfer PGD$_{\delta}$-20 attack at $\underline{\beta=10.0}$.}
\label{tab:TPGD_appendix_10}
\end{table}


\begin{table}[H]
\tiny
\centering
\begin{tabular}{c|c|c|c|c|c|c|c|c|c|c|c|c|c|c}
$\delta$ & Quant. & \multicolumn{7}{c}{$\epsilon$} \\
\cline{3-15}
 & after & $0$ & $1$ & $2$ & $3$ & $4$ & $5$ & $6$ & $7$ & $8$ & $9$ & $10$ & $11$ & $12$ \\
 & layer: & & & & & & & & & & & & & \\
\hline

2 & none & \cellcolor{gray!20}18.15 & \cellcolor{gray!20}71.84 & \cellcolor{gray!20}77.85 & \cellcolor{gray!20}79.01 & \cellcolor{gray!20}79.08 & \cellcolor{gray!20}78.75 & \cellcolor{gray!20}77.77 & \cellcolor{gray!20}77.05 & \cellcolor{gray!20}76.69 & \cellcolor{gray!20}75.57 & \cellcolor{gray!20}74.13 & \cellcolor{gray!20}72.28 & \cellcolor{gray!20}71.01  \\
 & conv0 & \cellcolor{blue!40}26.63& \cellcolor{blue!40}74.31& \cellcolor{blue!40}78.74& \cellcolor{blue!40}79.42& \cellcolor{blue!40}79.62& \cellcolor{blue!40}79.00& \cellcolor{red!6}77.75& \cellcolor{red!6}76.70& \cellcolor{red!6}76.40& \cellcolor{red!6}75.16& \cellcolor{red!6}73.27& \cellcolor{red!6}71.85& \cellcolor{red!6}70.34 \\
 & layer1 & \cellcolor{blue!7}18.34& \cellcolor{blue!11}72.53& \cellcolor{blue!16}78.21& \cellcolor{blue!27}79.29& \cellcolor{blue!9}79.21& \cellcolor{blue!24}78.90& \cellcolor{blue!32}77.89& \cellcolor{red!6}76.87& \cellcolor{red!6}76.63& \cellcolor{red!6}75.24& \cellcolor{red!6}73.98& \cellcolor{red!6}71.88& \cellcolor{red!6}70.71 \\
 & layer2 & \cellcolor{blue!7}18.44& \cellcolor{blue!7}72.22& \cellcolor{blue!7}77.88& \cellcolor{blue!13}79.15& \cellcolor{blue!10}79.22& \cellcolor{blue!14}78.84& \cellcolor{blue!37}77.91& \cellcolor{red!6}76.84& \cellcolor{blue!40}76.76& \cellcolor{red!6}75.55& \cellcolor{red!6}73.85& \cellcolor{red!6}71.98& \cellcolor{red!6}70.55 \\
 & layer3 & \cellcolor{blue!7}18.32& \cellcolor{blue!7}72.12& \cellcolor{blue!7}77.96& \cellcolor{blue!7}79.03& \cellcolor{blue!10}79.22& \cellcolor{blue!12}78.83& \cellcolor{blue!40}77.92& \cellcolor{blue!40}77.20& \cellcolor{blue!11}76.71& \cellcolor{blue!40}75.86& \cellcolor{blue!40}74.19& \cellcolor{blue!40}72.47& \cellcolor{blue!40}71.10 \\
\hline

\end{tabular}
\caption{Evaluation of quantization after four intermediate features of $\epsilon$-robust \underline{ResNet-18} models on the Transfer PGD$_{\delta}$-20 attack at $\underline{\beta=12.0}$.}
\label{tab:TPGD_appendix_12}
\end{table}


\begin{table}[H]
\tiny
\centering
\begin{tabular}{c|c|c|c|c|c|c|c|c}
$\delta$ & Quant. & \multicolumn{7}{c}{$\epsilon$} \\
\cline{3-9}
 & after & $0$ & $2$ & $4$ & $6$ & $8$ & $10$ & $12$ \\
 & layer: & & & & & & \\
\hline

2 & none & \cellcolor{gray!20}2.43 & \cellcolor{gray!20}81.77 & \cellcolor{gray!20}82.67 & \cellcolor{gray!20}81.58 & \cellcolor{gray!20}79.19 & \cellcolor{gray!20}76.23 & \cellcolor{gray!20}71.20  \\
 & init conv & \cellcolor{red!6}2.35& \cellcolor{red!6}81.63& \cellcolor{blue!40}82.76& \cellcolor{blue!40}81.60& \cellcolor{red!6}79.06& \cellcolor{red!6}75.95& \cellcolor{red!6}70.90 \\
 & layer[0] & \cellcolor{red!6}2.29& \cellcolor{blue!40}81.85& \cellcolor{red!6}82.60& \cellcolor{red!6}81.42& \cellcolor{red!6}78.92& \cellcolor{red!6}76.05& \cellcolor{red!6}70.92 \\
 & layer[1] & \cellcolor{red!6}2.40& \cellcolor{red!6}81.72& \cellcolor{blue!40}82.76& \cellcolor{red!6}81.54& \cellcolor{red!6}79.01& \cellcolor{red!6}75.99& \cellcolor{red!6}70.46 \\
 & layer[2] & \cellcolor{blue!40}2.47& \cellcolor{red!6}81.68& \cellcolor{blue!31}82.74& \cellcolor{red!6}81.45& \cellcolor{red!6}79.00& \cellcolor{red!6}75.98& \cellcolor{red!6}70.78 \\
\hline

\end{tabular}
\caption{Evaluation of quantization after four intermediate features of $\epsilon$-robust \underline{WideResNet-28-10} models on the Transfer PGD$_{\delta}$-20 attack for various $\epsilon$ and $\delta$ at $\underline{\beta=8.0}$. 
}
\label{tab:TPGD_wide_appendix}
\end{table}
\begin{table}[H]
\tiny
\centering
\begin{minipage}{.45\textwidth}
\begin{tabular}{c|c|c|c|c|c|c|c|c}
$\delta$ & Quant. & \multicolumn{7}{c}{$\epsilon$} \\
\cline{3-9}
 & after & $0$ & $2$ & $4$ & $6$ & $8$ & $10$ & $12$ \\
 & layer: & & & & & & \\
\hline

2 & none & \cellcolor{gray!20}2.43 & \cellcolor{gray!20}81.77 & \cellcolor{gray!20}82.67 & \cellcolor{gray!20}81.58 & \cellcolor{gray!20}79.19 & \cellcolor{gray!20}76.23 & \cellcolor{gray!20}71.20  \\
 & init conv & \cellcolor{blue!10}2.45& \cellcolor{red!6}81.65& \cellcolor{blue!7}82.69& \cellcolor{blue!40}81.60& \cellcolor{red!6}78.93& \cellcolor{red!6}75.53& \cellcolor{red!6}70.61 \\
 & layer[0] & \cellcolor{red!6}2.19& \cellcolor{blue!40}82.02& \cellcolor{red!6}82.60& \cellcolor{red!6}81.23& \cellcolor{red!6}78.52& \cellcolor{red!6}76.08& \cellcolor{red!6}70.52 \\
 & layer[1] & \cellcolor{red!6}2.41& \cellcolor{red!6}81.72& \cellcolor{blue!40}82.80& \cellcolor{red!6}81.33& \cellcolor{red!6}78.68& \cellcolor{red!6}75.66& \cellcolor{red!6}69.62 \\
 & layer[2] & \cellcolor{blue!40}2.51& \cellcolor{red!6}81.65& 82.67 & \cellcolor{red!6}81.22& \cellcolor{red!6}78.64& \cellcolor{red!6}75.63& \cellcolor{red!6}70.19 \\
\hline

\end{tabular}
\caption{Evaluation of quantization after four intermediate features of $\epsilon$-robust \underline{WideResNet-28-10} models on the Transfer PGD$_{\delta}$-20 attack at $\underline{\beta=4.0}$.}
\label{tab:TPGD_wide_appendix_4}
\end{minipage}
\hfill 
\begin{minipage}{.45\textwidth}

\begin{tabular}{c|c|c|c|c|c|c|c|c}
$\delta$ & Quant. & \multicolumn{7}{c}{$\epsilon$} \\
\cline{3-9}
 & after & $0$ & $2$ & $4$ & $6$ & $8$ & $10$ & $12$ \\
 & layer: & & & & & & \\
\hline

2 & none & \cellcolor{gray!20}2.43 & \cellcolor{gray!20}81.77 & \cellcolor{gray!20}82.67 & \cellcolor{gray!20}81.58 & \cellcolor{gray!20}79.19 & \cellcolor{gray!20}76.23 & \cellcolor{gray!20}71.20  \\
 & init conv & \cellcolor{red!6}2.37& \cellcolor{red!6}81.63& \cellcolor{blue!40}82.82& \cellcolor{red!6}81.55& \cellcolor{red!6}79.12& \cellcolor{red!6}75.80& \cellcolor{red!6}70.85 \\
 & layer[0] & \cellcolor{red!6}2.26& \cellcolor{blue!40}81.90& \cellcolor{red!6}82.55& \cellcolor{red!6}81.39& \cellcolor{red!6}78.75& \cellcolor{red!6}76.10& \cellcolor{red!6}70.89 \\
 & layer[1] & \cellcolor{red!6}2.40& \cellcolor{red!6}81.69& \cellcolor{blue!21}82.75& \cellcolor{red!6}81.44& \cellcolor{red!6}78.86& \cellcolor{red!6}75.97& \cellcolor{red!6}70.21 \\
 & layer[2] & \cellcolor{blue!40}2.50& \cellcolor{red!6}81.68& \cellcolor{blue!21}82.75& \cellcolor{red!6}81.40& \cellcolor{red!6}78.96& \cellcolor{red!6}75.82& \cellcolor{red!6}70.55 \\
\hline

\end{tabular}
\caption{Evaluation of quantization after four intermediate features of $\epsilon$-robust \underline{WideResNet-28-10} models on the Transfer PGD$_{\delta}$-20 attack at $\underline{\beta=6.0}$.}
\label{tab:TPGD_wide_appendix_6}
\end{minipage}
\end{table}


\begin{table}[H]
\tiny
\centering
\begin{minipage}{.45\textwidth}
\begin{tabular}{c|c|c|c|c|c|c|c|c}
$\delta$ & Quant. & \multicolumn{7}{c}{$\epsilon$} \\
\cline{3-9}
 & after & $0$ & $2$ & $4$ & $6$ & $8$ & $10$ & $12$ \\
 & layer: & & & & & & \\
\hline

2 & none & \cellcolor{gray!20}2.43 & \cellcolor{gray!20}81.77 & \cellcolor{gray!20}82.67 & \cellcolor{gray!20}81.58 & \cellcolor{gray!20}79.19 & \cellcolor{gray!20}76.23 & \cellcolor{gray!20}71.20  \\
 & init conv & \cellcolor{red!6}2.39& \cellcolor{red!6}81.61& \cellcolor{blue!40}82.74& \cellcolor{blue!40}81.59& \cellcolor{blue!40}79.20& \cellcolor{red!6}76.04& \cellcolor{red!6}71.04 \\
 & layer[0] & \cellcolor{red!6}2.31& \cellcolor{blue!40}81.80& \cellcolor{red!6}82.66& \cellcolor{red!6}81.46& \cellcolor{red!6}78.92& \cellcolor{red!6}76.10& \cellcolor{red!6}71.12 \\
 & layer[1] & \cellcolor{red!6}2.42& \cellcolor{red!6}81.75& \cellcolor{blue!40}82.74& \cellcolor{red!6}81.57& \cellcolor{red!6}79.11& \cellcolor{red!6}76.07& \cellcolor{red!6}70.70 \\
 & layer[2] & \cellcolor{blue!40}2.45& \cellcolor{red!6}81.72& \cellcolor{blue!34}82.73& \cellcolor{red!6}81.43& \cellcolor{red!6}79.02& \cellcolor{red!6}76.06& \cellcolor{red!6}70.86 \\
\hline

\end{tabular}
\caption{Evaluation of quantization after four intermediate features of $\epsilon$-robust \underline{WideResNet-28-10} models on the Transfer PGD$_{\delta}$-20 attack at $\underline{\beta=10.0}$.}
\label{tab:TPGD_wide_appendix_10}
\end{minipage}
\hfill 
\begin{minipage}{.45\textwidth}

\begin{tabular}{c|c|c|c|c|c|c|c|c}
$\delta$ & Quant. & \multicolumn{7}{c}{$\epsilon$} \\
\cline{3-9}
 & after & $0$ & $2$ & $4$ & $6$ & $8$ & $10$ & $12$ \\
 & layer: & & & & & & \\
\hline

2 & none & \cellcolor{gray!20}2.43 & \cellcolor{gray!20}81.77 & \cellcolor{gray!20}82.67 & \cellcolor{gray!20}81.58 & \cellcolor{gray!20}79.19 & \cellcolor{gray!20}76.23 & \cellcolor{gray!20}71.20  \\
 & init conv & \cellcolor{red!6}2.37& \cellcolor{red!6}81.64& \cellcolor{blue!40}82.78& \cellcolor{blue!40}81.60& \cellcolor{blue!40}79.24& \cellcolor{red!6}76.02& \cellcolor{red!6}71.06 \\
 & layer[0] & \cellcolor{red!6}2.36& \cellcolor{blue!40}81.84& \cellcolor{red!6}82.66& \cellcolor{red!6}81.46& \cellcolor{red!6}78.95& \cellcolor{red!6}76.02& 71.20  \\
 & layer[1] & 2.43 & \cellcolor{blue!17}81.80& \cellcolor{blue!14}82.71& 81.58 & \cellcolor{red!6}79.14& \cellcolor{red!6}76.05& \cellcolor{red!6}70.83 \\
 & layer[2] & \cellcolor{blue!40}2.44& \cellcolor{red!6}81.73& \cellcolor{blue!32}82.76& \cellcolor{red!6}81.44& \cellcolor{red!6}79.09& \cellcolor{red!6}76.03& \cellcolor{red!6}70.98 \\
\hline

\end{tabular}
\caption{Evaluation of quantization after four intermediate features of $\epsilon$-robust \underline{WideResNet-28-10} models on the Transfer PGD$_{\delta}$-20 attack at $\underline{\beta=12.0}$.}
\label{tab:TPGD_wide_appendix_12}
\end{minipage}
\end{table}

\subsection{Clean Accuracies After Quantization for Various $\beta$}
The clean accuracies after quantization with different scaling factors $\beta \in \{4, 6, 8, 10, 12\}$ is depicted below in Tables \ref{tab:quant_clean} and \ref{tab:quant_wide_clean}. We note that there is very minimal change in clean accuracies across values of $\epsilon$.
\begin{table}[H]
\tiny
\centering
\begin{tabular}{c|c|c|c|c|c|c|c|c|c|c|c|c|c|c}
$\beta$ & Quant. & \multicolumn{7}{c}{$\epsilon$} \\
\cline{3-15}
 & after & $0$ & $1$ & $2$ & $3$ & $4$ & $5$ & $6$ & $7$ & $8$ & $9$ & $10$ & $11$ & $12$ \\
 & layer: & & & & & & & & & & & & & \\
\hline

4 & none & 85.58 & 92.25 & 91.79 & 90.16 & 89.09 & 87.95 & 86.35 & 84.66 & 83.63 & 82.23 & 80.32 & 78.33 & 76.50  \\
 & conv0 & 65.40 & 85.60& 86.70 & 85.00 & 84.00 & 82.30 & 85.00 & 82.70 & 81.00 & 79.10 & 77.80 & 75.00 & 73.50  \\
 & layer1 & 84.50 & 91.50& 91.40 & 89.00 & 88.00 & 86.00 & 85.20 & 83.50 & 82.00 & 79.90 & 78.70 & 75.90 & 73.80  \\
 & layer2 & 84.50 & 91.00& 91.30 & 89.90 & 88.70 & 87.20 & 85.40 & 83.60 & 82.20 & 81.20 & 77.60 & 75.80 & 73.60  \\
 & layer3 & 85.30 & 92.00& 91.30 & 89.80 & 88.60 & 87.00 & 86.30 & 84.50 & 83.10 & 81.90 & 79.00 & 78.00 & 75.90  \\
 & layer4 & 85.30 & 92.30& 91.70 & 90.10 & 89.00 & 87.00 & 86.20 & 84.00 & 83.20 & 81.70 & 79.90 & 77.40 & 75.40  \\
\hline

6 & none & 85.58 & 92.25 & 91.79 & 90.16 & 89.09 & 87.95 & 86.35 & 84.66 & 83.63 & 82.23 & 80.32 & 78.33 & 76.50  \\
 & conv0 & 75.40 & 90.30 & 89.70 & 88.00 & 86.80& 85.10& 82.70& 81.60 & 80.00 & 78.70 & 75.50 & 73.10 & 72.00  \\
 & layer1 & 84.90 & 92.10 & 91.70 & 89.80 & 88.50& 87.20& 85.90& 84.10 & 83.00 & 81.00 & 79.40 & 76.80 & 75.20  \\
 & layer2 & 84.90 & 92.10 & 91.50 & 90.10 & 88.70& 87.00& 85.90& 84.00 & 82.70 & 81.70 & 78.90 & 77.00 & 75.10  \\
 & layer3 & 85.50 & 92.20 & 91.60 & 89.80 & 88.90& 88.00& 86.40& 84.60 & 83.20 & 82.00 & 79.90 & 78.20 & 76.20  \\
 & layer4 & 85.40 & 92.20 & 91.70 & 90.10 & 89.10& 87.80& 86.30& 84.00 & 83.30 & 82.00 & 80.00 & 77.80 & 76.00  \\
\hline

8 & none & 85.58 & 92.25 & 91.79 & 90.16 & 89.09 & 87.95 & 86.35 & 84.66 & 83.63 & 82.23 & 80.32 & 78.33 & 76.50  \\
 & conv0 & 79.90 & 91.30 & 90.70 & 88.70 & 87.70& 86.20 & 84.40& 82.60& 82.00 & 80.00 & 77.50 & 75.70 & 73.60  \\
 & layer1 & 85.10 & 92.00 & 91.70 & 89.80 & 88.60& 87.40 & 86.00& 84.40& 83.20 & 81.40 & 79.70 & 77.20 & 75.60  \\
 & layer2 & 85.10 & 92.20 & 91.50 & 90.10 & 88.80& 87.80 & 85.90& 84.30& 83.00 & 81.70 & 79.40 & 77.40 & 75.40  \\
 & layer3 & 85.50 & 92.20 & 91.70 & 90.00 & 89.00& 87.90 & 86.50& 84.70& 83.40 & 82.10 & 80.00 & 78.30 & 76.20  \\
 & layer4 & 85.40 & 92.20 & 91.70 & 90.10 & 89.10& 87.90 & 86.40& 84.50& 83.40 & 82.00 & 80.10 & 78.00 & 76.10  \\
\hline

10 & none & 85.58 & 92.25 & 91.79 & 90.16 & 89.09 & 87.95 & 86.35 & 84.66 & 83.63 & 82.23 & 80.32 & 78.33 & 76.50  \\
 & conv0 & 81.40 & 91.00& 91.10 & 89.20 & 88.10& 86.00 & 85.10& 83.30& 82.30 & 80.00 & 78.30 & 76.50 & 74.50  \\
 & layer1 & 85.20 & 92.30& 91.70 & 89.00 & 88.70& 87.50 & 86.00& 84.40& 83.00 & 81.60 & 79.80 & 77.40 & 75.90  \\
 & layer2 & 85.20 & 92.20& 91.70 & 90.00 & 88.90& 87.70 & 86.00& 84.40& 83.20 & 81.80 & 79.00 & 77.50 & 75.00  \\
 & layer3 & 85.00 & 92.20& 91.70 & 90.00 & 89.00& 87.90 & 86.50& 84.70& 83.00 & 82.20 & 80.20 & 78.20 & 76.30  \\
 & layer4 & 85.40 & 92.20& 91.70 & 90.10 & 89.10& 87.90 & 86.40& 84.60& 83.40 & 82.00 & 80.20 & 78.00 & 76.00  \\
\hline

12 & none & 85.58 & 92.25 & 91.79 & 90.16 & 89.09 & 87.95 & 86.35 & 84.66 & 83.63 & 82.23 & 80.32 & 78.33 & 76.50  \\
 & conv0 & 82.60 & 91.90 & 91.00& 89.40& 88.50& 86.90 & 85.20& 83.60& 82.50 & 81.00& 78.00 & 76.80 & 74.90  \\
 & layer1 & 85.30 & 92.20 & 91.70& 89.90& 88.80& 87.00 & 86.20& 84.50& 83.20 & 81.70& 79.90 & 77.60 & 76.00  \\
 & layer2 & 85.20 & 92.20 & 91.70& 90.20& 88.90& 87.80 & 86.10& 84.50& 83.30 & 81.90& 79.70 & 77.70 & 75.80  \\
 & layer3 & 85.50 & 92.20 & 91.80& 90.00& 89.00& 87.90 & 86.60& 84.70& 83.50 & 82.30& 80.10 & 78.30 & 76.00  \\
 & layer4 & 85.40 & 92.20 & 91.70& 90.10& 89.10& 87.90 & 86.30& 84.60& 83.00 & 82.00& 80.20 & 78.10 & 76.30  \\
\hline

\end{tabular}
\caption{Clean accuracy evaluation of quantization at various $\beta\in\{4, 6, 8, 10, 12\}$ after four intermediate features of $\epsilon$-robust \underline{ResNet18} models. 
}
\label{tab:quant_clean}
\end{table}
\begin{table}[H]
\tiny
\centering
\begin{tabular}{c|c|c|c|c|c|c|c|c}
$\beta$ & Quant. & \multicolumn{7}{c}{$\epsilon$} \\
\cline{3-9}
 & after & $0$ & $2$ & $4$ & $6$ & $8$ & $10$ & $12$ \\
 & layer: & & & & & & \\
\hline

4 & none & 85.45 & 93.41 & 91.13 & 88.66 & 85.23 & 81.51 & 75.83  \\
 & init conv & 84.00& 93.30 & 90.90 & 88.30 & 84.80 & 81.00 & 75.50  \\
 & layer[0] & 85.60& 93.10 & 90.90 & 88.10 & 84.60 & 81.30 & 75.10  \\
 & layer[1] & 85.10& 93.30 & 91.00 & 88.40 & 84.80 & 80.70 & 74.30  \\
 & layer[2] & 85.10& 93.30 & 91.10 & 88.40 & 84.70 & 80.70 & 74.80  \\
\hline

6 & none & 85.45 & 93.41 & 91.13 & 88.66 & 85.23 & 81.51 & 75.83  \\
 & init conv & 84.90& 93.30& 91.00 & 88.40 & 85.00 & 81.30& 75.70  \\
 & layer[0] & 85.70& 93.20& 91.00 & 88.30 & 85.00 & 81.60& 75.50  \\
 & layer[1] & 85.30& 93.50& 91.00 & 88.50 & 84.90 & 80.90& 75.00  \\
 & layer[2] & 85.10& 93.30& 91.10 & 88.50 & 84.90 & 81.00& 75.10  \\
\hline

8 & none & 85.45 & 93.41 & 91.13 & 88.66 & 85.23 & 81.51 & 75.83  \\
 & init conv & 84.90& 93.30 & 91.00& 88.40 & 85.20 & 81.40 & 75.80  \\
 & layer[0] & 85.60& 93.30 & 91.00& 88.40 & 85.10 & 81.50 & 75.60  \\
 & layer[1] & 85.40& 93.00 & 91.00& 88.60 & 85.00 & 81.00 & 75.20  \\
 & layer[2] & 85.10& 93.30 & 91.20& 88.50 & 85.00 & 81.20 & 75.30  \\
\hline

10 & none & 85.45 & 93.41 & 91.13 & 88.66 & 85.23 & 81.51 & 75.83  \\
 & init conv & 85.20& 93.30 & 91.00 & 88.50 & 85.10 & 81.30 & 75.80  \\
 & layer[0] & 85.50& 93.30 & 91.00 & 88.40 & 85.10 & 81.50 & 75.70  \\
 & layer[1] & 85.00& 93.40 & 91.00 & 88.60 & 85.10 & 81.20 & 75.30  \\
 & layer[2] & 85.20& 93.30 & 91.10 & 88.50 & 85.10 & 81.30 & 75.40  \\
\hline

12 & none & 85.45 & 93.41 & 91.13 & 88.66 & 85.23 & 81.51 & 75.83  \\
 & init conv & 85.20& 93.30 & 91.00& 88.50 & 85.20 & 81.30 & 75.80  \\
 & layer[0] & 85.50& 93.40 & 91.00& 88.00 & 85.10 & 81.50 & 75.70  \\
 & layer[1] & 85.40& 93.40 & 91.00& 88.00 & 85.10 & 81.30 & 75.30  \\
 & layer[2] & 85.30& 93.30 & 91.20& 88.60 & 85.10 & 81.30 & 75.50  \\
\hline

\end{tabular}
\caption{Clean accuracy evaluation of quantization at various $\beta\in\{4, 6, 8, 10, 12\}$ after four intermediate features of $\epsilon$-robust \underline{WideResNet-28-10} models. 
}
\label{tab:quant_wide_clean}
\end{table}



\section{AT and Norm of CNN Kernels}\label{app:change}
\begin{figure}[H]
    \centering
    \begin{subfigure}{0.44\textwidth}
        \includegraphics[width=\linewidth]{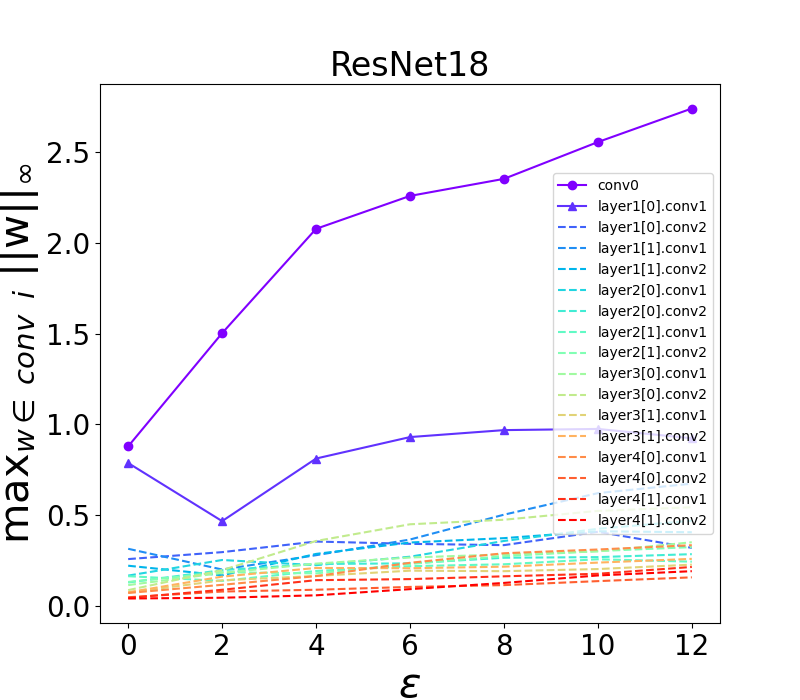}
    \end{subfigure}
    \begin{subfigure}{0.55\textwidth}
        \includegraphics[width=\linewidth]{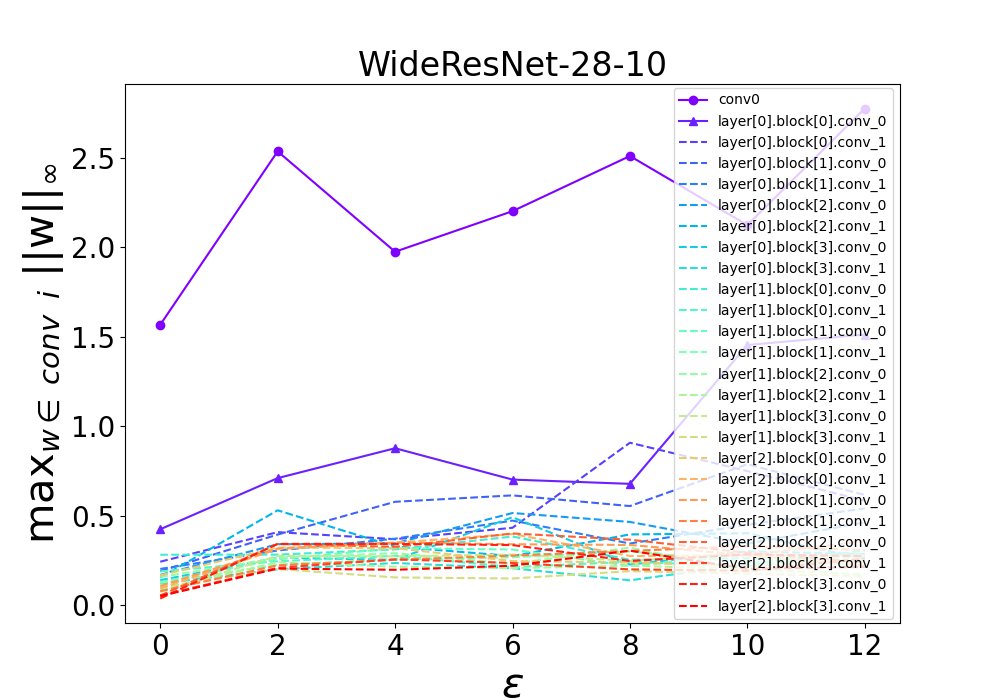}
    \end{subfigure}
    \caption{Plot of maximum of $L_{\infty}$ norms of filters in various layers for increasingly $\epsilon$-robust ResNet18 \emph{(left)} and WideResNet-28-10 \emph{(right)} models.}
    \label{fig:appendix_change}
\end{figure}

We analyze the worst case performance by plotting (for convolution layer $i$), $\max_{w \in \; \text{conv} \; i} ||w||_{\infty}$ for various layers of increasing $\epsilon$-robust models in Figure \ref{fig:appendix_change}.
We note a large separation between the first two layers and other layers (as previously observed in the average case in Section \ref{sec:change}) but also observe a monotonic increase for the first two curves of ResNet18 but not WideResNet-28-10.

\section{Training with Larger Perturbations and Common-Corruptions}\label{app:OOD_all}
We present the accuracy of non-robust and various $\epsilon$-robust models for each corruption with ResNet18 models, and WideResNet-18 models in Tables \ref{tab:many_corruptions_appendix}, and \ref{tab:many_corruptions_wide_appendix} respectively.
\begin{table}[H]
    \scriptsize
    \centering
    \begin{tabular}{l|c|c|c|c|c|c|c|c|c|c|c|c|c}

corruption & non-robust & 1 & 2 & 3 & 4 & 5 & 6 & 7 & 8 & 9 & 10 & 11 & 12 \\
\hline
gaussian noise & 51.33 & 80.59 & \underline{84.09} & \textbf{84.59} & 83.75 & 82.65 & 81.63 & 80.28 & 79.34 & 78.25 & 76.66 & 74.49 & 73.2 \\
shot noise & 60.0 & 84.34 & \textbf{86.06} & \underline{85.92} & 84.91 & 83.63 & 82.56 & 81.06 & 80.19 & 78.93 & 77.45 & 75.23 & 73.82 \\
impulse noise & 55.82 & 68.8 & 73.3 & \textbf{76.08} & 75.34 & \underline{75.94} & 74.58 & 74.08 & 73.7 & 73.61 & 72.97 & 71.26 & 70.59 \\
defocus blur & 71.58 & \underline{84.64} & \textbf{85.02} & 84.47 & 83.62 & 82.74 & 81.61 & 80.01 & 79.27 & 77.86 & 76.39 & 74.43 & 72.94 \\
glass blur & 46.46 & 75.54 & 79.53 & \underline{80.4} & \textbf{80.51} & 80.21 & 79.58 & 78.37 & 77.54 & 76.5 & 74.83 & 72.86 & 71.52 \\
motion blur & 61.78 & 79.65 & \underline{79.89} & \textbf{80.32} & 79.1 & 78.71 & 77.6 & 76.13 & 75.61 & 74.3 & 72.58 & 71.22 & 69.85 \\
zoom blur & 68.7 & \underline{83.51} & \textbf{83.73} & 83.42 & 82.36 & 81.68 & 80.82 & 79.35 & 78.58 & 77.08 & 75.57 & 73.98 & 72.43 \\
snow & 68.29 & \underline{84.94} & \textbf{85.02} & 83.68 & 82.87 & 81.52 & 80.31 & 78.96 & 77.9 & 76.26 & 75.11 & 72.95 & 71.35 \\
frost & 67.53 & \textbf{85.79} & \underline{84.86} & 82.66 & 80.85 & 78.8 & 76.96 & 74.93 & 73.19 & 70.98 & 69.1 & 66.9 & 64.67 \\
fog & \underline{73.15} & \textbf{75.59} & 70.56 & 67.13 & 65.01 & 62.91 & 61.2 & 59.29 & 58.49 & 57.12 & 55.44 & 54.54 & 53.14 \\
brightness & 81.99 & \textbf{90.62} & \underline{89.63} & 87.53 & 86.17 & 84.33 & 82.73 & 81.21 & 80.03 & 78.05 & 76.5 & 74.47 & 72.63 \\
contrast & \underline{57.68} & \textbf{59.15} & 53.91 & 51.19 & 48.77 & 46.44 & 45.02 & 43.04 & 42.46 & 41.43 & 40.42 & 40.08 & 39.23 \\
elastic transform & 73.0 & \textbf{85.19} & \underline{84.86} & 84.0 & 82.94 & 81.88 & 80.72 & 79.24 & 78.29 & 76.8 & 74.92 & 73.23 & 71.56 \\
pixelate & 65.45 & \underline{88.42} & \textbf{88.88} & 87.81 & 86.89 & 85.79 & 84.36 & 82.64 & 81.7 & 80.3 & 78.62 & 76.49 & 74.92 \\
jpeg compression & 72.0 & \underline{88.94} & \textbf{89.13} & 87.99 & 86.73 & 85.7 & 84.33 & 82.58 & 81.79 & 80.44 & 78.59 & 76.47 & 75.14 \\
speckle noise & 61.67 & 83.76 & \underline{85.45} & \textbf{85.55} & 84.63 & 83.38 & 81.98 & 80.8 & 79.92 & 78.73 & 77.28 & 74.95 & 73.6 \\
gaussian blur & 65.23 & 80.94 & \underline{81.75} & \textbf{81.94} & 81.08 & 80.51 & 79.38 & 77.93 & 77.16 & 75.99 & 74.46 & 72.73 & 71.39 \\
spatter & 73.13 & \textbf{85.72} & \underline{85.25} & 84.29 & 83.59 & 82.42 & 81.34 & 80.2 & 79.16 & 78.0 & 76.31 & 74.54 & 73.06 \\
saturate & 79.28 & \textbf{88.04} & \underline{86.59} & 85.38 & 84.42 & 83.06 & 82.04 & 80.27 & 79.49 & 78.07 & 76.52 & 74.68 & 72.89 \\
\hline
\hline
Avg. & 66.0 & \underline{81.8} & \textbf{81.97} & 81.28 & 80.19 & 79.07 & 77.83 & 76.33 & 75.46 & 74.14 & 72.62 & 70.82 & 69.37 

    \end{tabular}
    \caption{Accuracy of $\epsilon$-robust models (from $\epsilon=$1 to 12) on various common corruptions. The bold and underlined values denote the largest and second largest value in each row (\textit{i.e.}, for each type of corruption).}
    \label{tab:many_corruptions_appendix}
\end{table}
\begin{table}[H]
    \scriptsize
    \centering
    \begin{tabular}{l|c|c|c|c|c|c|c|c|c|c|c|c|c}

corruption & non-robust & 2 & 4 & 6 & 8 & 10 & 12 \\
\hline
gaussian noise & 28.42 & \underline{85.44} & \textbf{86.26} & 83.55 & 80.59 & 77.55 & 72.5 \\
shot noise & 38.11 & \textbf{87.43} & \underline{87.29} & 84.8 & 81.71 & 78.4 & 73.33 \\
impulse noise & 48.51 & 74.52 & \textbf{78.34} & \underline{77.15} & 75.37 & 74.28 & 71.29 \\
defocus blur & 67.26 & \textbf{86.73} & \underline{85.1} & 83.0 & 80.14 & 76.93 & 71.49 \\
glass blur & 34.11 & 79.06 & \textbf{81.64} & \underline{80.07} & 77.81 & 74.97 & 69.8 \\
motion blur & 59.05 & \textbf{81.83} & \underline{80.76} & 78.73 & 75.93 & 73.09 & 68.42 \\
zoom blur & 58.12 & \textbf{85.6} & \underline{83.69} & 82.04 & 79.0 & 75.9 & 70.56 \\
snow & 62.43 & \textbf{87.52} & \underline{86.19} & 83.11 & 79.98 & 76.56 & 70.91 \\
frost & 56.36 & \textbf{87.34} & \underline{84.79} & 80.14 & 75.9 & 70.26 & 63.72 \\
fog & \underline{72.36} & \textbf{74.35} & 67.98 & 62.79 & 59.07 & 56.41 & 52.5 \\
brightness & 83.06 & \textbf{91.84} & \underline{89.38} & 85.76 & 82.05 & 78.23 & 71.71 \\
contrast & \underline{55.34} & \textbf{58.58} & 50.42 & 45.36 & 41.95 & 40.42 & 38.19 \\
elastic transform & 67.85 & \textbf{86.52} & \underline{84.45} & 82.01 & 78.86 & 75.52 & 70.18 \\
pixelate & 61.24 & \textbf{90.03} & \underline{88.61} & 85.99 & 83.06 & 79.5 & 73.84 \\
jpeg compression & 65.8 & \textbf{90.47} & \underline{88.54} & 86.0 & 82.89 & 79.44 & 73.9 \\
speckle noise & 40.73 & \textbf{86.91} & \underline{86.75} & 84.58 & 81.57 & 78.27 & 73.12 \\
gaussian blur & 55.78 & \textbf{83.45} & \underline{82.4} & 80.8 & 77.91 & 74.88 & 69.7 \\
spatter & 75.03 & \textbf{88.26} & \underline{86.76} & 84.12 & 81.11 & 77.79 & 72.64 \\
saturate & 81.47 & \textbf{90.04} & \underline{87.96} & 85.07 & 82.02 & 78.65 & 73.28 \\
\hline
\hline
Avg. & 58.48 & \textbf{84.0} & \underline{82.49} & 79.74 & 76.68 & 73.53 & 68.48

    \end{tabular}
    \caption{Accuracy of $\epsilon$-robust WideResNet-18 models ($\epsilon=2,4,6,8,10,12$) on various common corruptions. The bold and underlined values denote the largest and second largest value in each row (\textit{i.e.}, for each type of corruption).}
    \label{tab:many_corruptions_wide_appendix}
\end{table}

\end{document}